\documentclass[preprint,12pt]{elsarticle}

\usepackage{amsmath,amssymb,amsfonts}
\usepackage{graphicx}
\usepackage[hidelinks]{hyperref}
\usepackage{booktabs}
\usepackage{pifont}
\usepackage{multirow}
\usepackage{xcolor}
\usepackage{adjustbox}
\usepackage{subcaption}

\newcommand{\cmark}{\ding{51}}%
\newcommand{\xmark}{\ding{55}}%
\newcommand{\sieno}{\ding{81}}%
\usepackage{wrapfig}
\usepackage{lscape}
\usepackage{rotating}
\usepackage{float}

\usepackage[]{tcolorbox}        

\definecolor{mygreen}{RGB}{32, 183, 8}
\definecolor{myred}{RGB}{255, 92, 51}
\definecolor{myblue}{RGB}{102, 140, 255}
\definecolor{myorange}{RGB}{252, 139, 33}

\tcbset{
    colback = white,
    before skip = 0.2cm,    
    after skip = 0.5cm      
}                           

\newtcolorbox{boxA}{
    boxrule = 0.2pt,
    colframe = black,
    colback=yellow!10!white
}

\journal{Journal of Artificial Intelligence in Medicine}

\begin{document}

\begin{frontmatter}



\title{Seamless Monitoring of Stress Levels Leveraging a Foundational  Model for Time Sequences} 


\author[sapienza]{Davide Gabrielli}

\author[sapienza,tum]{Bardh Prenkaj\fnref{supervision}}
\author[sapienza]{Paola Velardi\fnref{supervision}\corref{mycorrespondingauthor}}

\affiliation[sapienza]{
    organization={Sapienza University of Rome},
    addressline={Via Salaria 113},
    city={Rome},
    postcode={00198},
    country={Italy}
}

\affiliation[tum]{
    organization={Technical University of Munich},
    addressline={Richard-Wagner-Str. 1},
    city={Munich},
    postcode={80333},
    country={Germany}
}

\cortext[mycorrespondingauthor]{Corresponding author.}
\ead{velardi@di.uniroma1.it}

\fntext[supervision]{Equal supervision contribution.}

\begin{abstract}
Monitoring the stress level in patients with neurodegenerative diseases can help manage symptoms, improve patient's quality of life, and provide insight into disease progression. In the literature, ECG, actigraphy, speech, voice, and facial analysis have proven effective at detecting patients' emotions. On the other hand, these tools are invasive and do not integrate smoothly into the patient's daily life. HRV has also been proven to effectively indicate stress conditions, especially in combination with other signals. However, when HRV is derived from less invasive devices than the ECG, like wristbands and smartwatches, the quality of measurements significantly degrades. This paper presents a methodology for stress detection from a wristband based on a universal model for time series, UniTS, which we finetuned for the task and equipped with explainability features. We cast the problem as anomaly detection rather than classification to favor model adaptation to individual patients and allow the clinician to maintain greater control over the system's predictions. We demonstrate that our proposed model considerably surpasses 12 top-performing methods on three benchmark datasets. Furthermore, unlike other state-of-the-art systems, UniTS enables seamless monitoring, as it shows comparable performance when using signals from invasive or lightweight devices.\end{abstract}




\end{frontmatter}



\section{Introduction}\label{sec:introduction}

High stress levels can worsen the symptoms of neurodegenerative diseases such as Alzheimer's and Parkinson's~\cite{cells12232726}. In these patients, stress is generated by mood changes such as depression and apathy, confinement at home, and other factors such as social isolation, uncertainty about the evolution of the disease, and the economic, health, personal, and family situation~\cite{mentalhealthplosone}~\cite{heilman2023emotional}.  Analyzing stress levels can provide insight into disease progression and help healthcare providers adjust treatment plans accordingly. Managing stress can also improve patients' quality of life and help to inform new therapeutic strategies.

AI-based methods for stress detection are motivated by the need for early, timely, and personalized stress management solutions. These methods leverage cutting-edge machine learning algorithms to provide objective, real-time monitoring and analysis, significantly contributing to mental health and well-being (see, among others,~\cite{realtimementalstressdetection}).

Among the numerous techniques presented in the literature, those based on data extracted from sensor devices are particularly promising due to several advantages. One of the key benefits of using sensor-based methods is their ability to provide real-time data, which is crucial for timely interventions and stress management. Secondly, they protect privacy more than solutions based on behavioral signals such as speech, gestures, and facial expressions~\cite{facespeechtext}, a particularly relevant aspect for elderly and frail people (see~\cite{eledrprivacy}, among others). Third, they may support continuous monitoring, enabling more convenient home care solutions for patients and caregivers.

In the literature, various sensor types are used to detect stress, leveraging different physiological signals. As surveyed in ~\cite{reviewstress24},  commonly used devices are ECG and EDA, EEG, skin temperature, respiratory and pressure sensors, and activity sensors like accelerometers, gyroscopes, and RFID technology. Among these devices, only a few are well-suited for seamless monitoring due to invasiveness, lack of autonomy for patients, and high costs. We define \textit{seamless} monitoring as the real-time collection of signals that is i) without time limits, ii) uninterrupted, and iii) does not require explicit actions by patients and doctors.\footnote{In this sense, ECG is not seamless because it can usually last for 24-48 hours, and must be activated by the doctor or patient, depending upon the device type.} In other terms, while continuous monitoring only ensures that data is captured without interruption, seamless monitoring ensures that this process integrates smoothly into the patient's daily life.
To address this challenge, which is essential to facilitate the widespread adoption of monitoring technologies for stress detection and healthcare applications in general, we propose a methodology providing the following contributions:

\begin{enumerate}
    \item To detect stress signals from sensor data, we leverage for the first time a universal model for time sequences, UniTS~\cite{gao2024units}, achieving 22\% superior performance against the second best performing -- i.e., DAGMM~\cite{zong2018deep} -- among 12 compared state-of-the-art (SoTA) methods on three publicly available datasets;  
    \item The proposed model allows us to obtain, using data from lightweight devices, performances comparable to those obtained from more invasive devices, such as  ECG, thus enabling seamless monitoring; 
    \item We cast the problem of stress detection as anomaly detection rather than classification, as the majority of works in the literature do (see~\autoref{tab:sota}). Anomaly detection typically involves establishing a baseline of normality rather than learning boundaries between multiple classes. The former is a more straightforward process since any deviation from normal behavior can be visualized and investigated for specific reasons. This way, doctors can more easily understand and trust the system's predictions, enhancing their ability to provide high-quality care.
    \item We integrate an LLM to translate anomalies in human-readable explanations based on personalized prompting~\cite{DBLP:journals/corr/abs-2402-18180} to support healthcare personnel in interpreting the detections UniTS makes. 
    \item We release our code publicly with all the model weights to encourage reproducibility from future researchers.\footnote{\url{https://github.com/davegabe/Wearable-Stress-Monitor}}
\end{enumerate}

\section{Related Work}
Our work relates to AI-based emotion monitoring from sensor data and anomaly detection from time series. 

\begin{table}[!h]
\centering
\caption{Organization of the literature on ML methods for mood monitoring from physiological data.}
\label{tab:sota}
\resizebox{\columnwidth}{!}{%
\begin{tabular}{@{}ccccccc@{}}
\toprule
                  & Signals Used         & Application & Device Type    & Model Type   & Task & Year \\ \midrule
\cite{jiaqi2023uaed}          & ECG                            & S & ECG, Wearable & Conv. Gaussian Mixture VAE              & AD & 2023 \\
\cite{al2023stress}           & Temp., Humidity (Sweat), Steps & S & Wristband    & Stacked Ens. with GB                    & C  & 2023 \\
\cite{can2019continuous}      & HR, GSR, Acc         & S           & Smartwatch     & MLP          & C    & 2019 \\
\cite{ghaderi2015stress}      & Resp., GSR, HR, EMG  & S           & \sieno         & SVM, kNN     & C    & 2015 \\
\cite{gjoreski2017monitoring} & BVP, HR, ST, GSR, Resp.        & S & Smartwatch   & RF                                      & C  & 2017 \\
\cite{sano2013stress}         & Steps, Acc           & S           & Wristband      & SVM, kNN     & C    & 2013 \\
\cite{kurniawan2013stress}    & GSR (+ Speech)       & S           & GSR Device     & SVM          & C    & 2013 \\
\cite{ahn2019novel}           & EEG, ECG             & S           & Wearable       & SVM          & C    & 2019 \\ \midrule
\cite{bulagang2021multiclass} & HR                   & E           & Wristband      & SVM, kNN, RF & C    & 2021 \\
\cite{guo2016heart}           & HRV (from ECG)       & E           & ECG, Wearable   & SVM          & C    & 2016 \\
\cite{hakim2018emotion}       & SpO2, Pulse Rate     & E           & Pulse Oximetry & SVM          & C    & 2018 \\
\cite{kim2008emotion}         & ECM, ECG, GSR, Resp. & E           & \sieno         & pLDA         & C    & 2008 \\
\cite{lin2021advanced}        & HR, BP                         & E & Smartwatch   & Fuzzy Petri Net~\cite{cardoso1996fuzzy} & C  & 2021 \\
$[\text{us}]$               & HRV, HR              & S + E       & Wristband, ECG & Transformer  & AD   & 2024 \\ \bottomrule
\end{tabular}%
}
\end{table}
\subsection{Emotion Monitoring and Stress Detection} 

\autoref{tab:sota} summarizes the key details regarding research on emotion recognition through physiological signals, providing a qualitative comparison with the solution proposed in this paper. (1) \textit{``Signals Used''} lists the physiological signals employed. Examples are Electrocardiogram (ECG), Heart Rate (HR), Galvanic Skin Response (GSR), Respiration (Resp.), and Electromyogram (EMG), among others. The choice of signals is crucial as it directly affects the accuracy and reliability of the detection system. Different signals provide varied information about an individual's physiological state, impacting the performance of the employed models. (2) \textit{``Application''} specifies whether the study focuses on stress (S) or emotion (E) detection, whereas the second considers nuanced mental conditions. (3) \textit{``Device Type''} specifies the type of device used to collect the physiological signals, such as wearables, smartwatches, wristbands, GSR devices, or pulse oximetry. The device type influences the monitoring system's accuracy, practicality, comfort, and user acceptance. Only some devices allow for seamless monitoring in real-world settings, which is especially valuable for frail people. (4) \textit{``Model Type''}  lists the machine learning or statistical models used for classification tasks. Notice how, except~\cite{al2023stress} and~\cite{jiaqi2023uaed}, all works rely on simple off-the-shelf machine learning models (SVM, kNN, and pLDA\footnote{Extended Linear Discriminant Analysis.}), which might not be suitable to detect emotions or stress levels that are not in the training data distribution, and therefore have limited capacity for generalization but also for personalization. In~\cite{al2023stress}, an ensemble mechanism over Gradient Boosts (GB) is applied to classify stress levels by leveraging the expertise of multiple models. In this case, a linear regression meta-model is applied to the output of the GB base models to predict the outcome. In ~\cite{jiaqi2023uaed}, a clustering algorithm based on a Gaussian mixture
convolutional variational autoencoder is employed to identify
normal emotional patterns and detect anomalies. To the best of our knowledge, we are the first to use a state-of-the-art foundational model for time-series to tackle stress and abnormal emotion detection. (5) \textit{``Task''}  indicates whether the model's task is classification (C) or anomaly detection (AD). Most works tackle stress and emotion monitoring as classification problems, categorizing states into predefined classes. The only exception is UAED~\cite{jiaqi2023uaed}, which performs an anomaly detection task. 
We argue that approaching stress monitoring as an anomaly detection problem is more principled and inherently more explainable than doing fine-grained emotion classification.

\subsection{Anomaly Detection in Time Series}\label{sec:ad_in_time_series} 

Following~\cite{jiaqi2023uaed}, we cast stress monitoring as an anomaly detection problem. We identified five methods proposed in the literature for anomaly detection in time series. Distance-based outlier detectors consider the distance of a point from its k-nearest neighbors. Density-based methods (among others~\cite{WangCNLCT21, su2019robust}) consider the density of the point and its neighbors. Prediction-based methods (among others,~\cite{benkabou2021local}) calculate the difference between the predicted and true values to detect anomalies. Reconstruction-based methods (among others~\cite{flaborea2023we}) compare the input signal and the reconstructed one in the output layer, typically using autoencoders. These methods assume anomalies are difficult to reconstruct and lost when the signal is mapped to lower dimensions, so a higher reconstruction error indicates a higher anomaly score. 

GAN-based anomaly detection methods have gained popularity recently. MAD-GAN~\cite{Li2019MADGANMA} combines the discriminator output with reconstruction error to detect anomalies in multivariate time series. TadGAN~\cite{geiger2020tadgan} uses a cycle-consistent GAN architecture with an encoder-decoder generator and proposes several ways to compute reconstruction error combined with critic outputs. HypAD~\cite{flaborea2023we} proposes using hyperbolic uncertainty for anomaly detection. TranAD~\cite{tuli2022tranad} is a transformer-based model that uses adversarial training and self-conditioning to capture broader temporal patterns effectively. USAD~\cite{audibert2020usad} combines adversarial training with an autoencoder, amplifying reconstruction errors for anomalies while ensuring robust learning.

Gaussian Mixture Model (GMM) methods include DAGMM~\cite{zong2018deep}, which integrates a deep autoencoder with a GMM for joint optimization, enhancing the alignment of dimensionality reduction and density estimation for unsupervised anomaly detection. Zhu et al.~\cite{jiaqi2023uaed} utilize a 2D Convolutional Variational Autoencoder with GMM latent distributions, transforming ECG signals into image-like formats for anomaly detection.

Graph-based methods have also emerged as viable approaches for time-series anomaly detection. MTAD-GAT~\cite{zhao2020multivariate} employs Graph Attention Networks (GAT) to model inter-feature and temporal dependencies, optimizing forecasting and reconstruction tasks jointly. GDN~\cite{deng2021graph} learns sensor relationships via graph structure learning and uses attention mechanisms for detecting deviations from learned patterns.

Among other deep learning-based approaches, OmniAnomaly~\cite{su2019robust} employs stochastic latent variables to model normal patterns in multivariate time series. CAE-M~\cite{zhang2021unsupervised} combines a convolutional autoencoder with a memory network for robust anomaly detection, incorporating a Maximum Mean Discrepancy (MMD) penalty to reduce noise impact. MSCRED~\cite{zhang2019deep} captures multi-scale system status using signature matrices and applies convolutional LSTM to model temporal dependencies, diagnosing anomalies based on inter-sensor correlations. LSTM-NDT~\cite{hundman2018detecting} utilizes LSTMs to forecast telemetry values with a dynamic thresholding mechanism for anomaly detection.

In this paper, we employ for the first time a foundation multi-task model specialized for time series, UniTS~\cite{gao2024units}, showing that it considerably surpasses best-performing anomaly detection models, including GAN, in the task of stress detection from physiological signals.

\section{Method}
\label{method}

UniTS~\cite{gao2024units} is the first foundation model\footnote{Foundation models are pre-trained on broad data at scale and can be adapted to a wide range of downstream tasks. } for multivariate time series, designed to perform in a unified way various tasks such as forecasting, imputation, classification, and anomaly detection. It employs self-attention to capture relationships across both the sequence and variable dimensions. This is particularly beneficial in our multi-sensor domain since, if one variable has an anomaly, the model can assess how this anomaly correlates with the patterns observed in other variables, thereby improving detection accuracy. Furthermore, UniTS is designed to cohesively integrate signals of different types and lengths. To achieve this, it employs multi-scale processing techniques to handle time-series data of different lengths, which involves processing the data at various temporal resolutions.

\subsection{Problem Definition and Preliminaries}
\label{sub: problemdefinition}
Anomaly detection in multivariate time series~\cite{audibert2020usad,flaborea2023we,prenkaj2023unsupervised} is the process of identifying observations within a sequence of time-dependent multivariate data that deviate significantly from the learned patterns of normal behavior. Formally, consider a multivariate time series \(\mathbf{X} = \{\mathbf{x}_1, \mathbf{x}_2, \ldots, \mathbf{x}_T\}\), where \(\mathbf{x}_t \in \mathbb{R}^n\) represents the \(n\)-dimensional observation at time \(t\), and \(T\) is the total number of time points. The objective is to develop a model \(f\) based on a historical subset of data \(\mathbf{X}_{train} = \{\mathbf{x}_1, \mathbf{x}_2, \ldots, \mathbf{x}_t\}\) for \(t < T\), that captures the normal behavior of the series.

For each time point \(t\), an anomaly score \(S_t = g(\mathbf{x}_t, f(\mathbf{X}_{train}))\) is computed, where \(g\) measures the deviation of \(\mathbf{x}_t\) from the expected behavior defined by \(f\). An observation \(\mathbf{x}_t\) is classified as abnormal if its anomaly score \(S_t\) exceeds a predefined threshold \(\tau\). Thus, the set of detected anomalies is given by \(\mathbf{A} = \{\mathbf{x}_t \mid S_t > \tau, t \in \{1, 2, \ldots, T\}\}\). 

According to~\cite{audibert2020usad}, to model the dependence between a current time point and previous ones, we can define a time window \( W_t \) of length \( K \) at a given time \( t \) as follows:
\begin{equation}
 W_t = \{\mathbf{x}_{t-K+1}, \ldots, \mathbf{x}_{t-1}, \mathbf{x}_t\}.
\end{equation}
It is then possible to transform the original time series \(\mathbf{X} = \{\mathbf{x}_1, \mathbf{x}_2, \ldots, \mathbf{x}_T\}\) into a sequence of windows \(\mathbf{W} = \{W_1, \ldots, W_T\}\) to be used as training input. Given a binary variable \( y \in \{0, 1\} \), the goal of our anomaly detection problem is to assign to an unseen window \(\widehat{W}_t, t > T\), a label \( y_t \) to indicate a detected anomaly at time \( t \), i.e., \( y_t = 1 \) for an anomaly or \( y_t = 0 \) for normal behavior, based on the window’s anomaly score. For the sake of simplicity and without loss of generality, we will use \( W \) to denote a training input window and \(\widehat{W}\) to denote an unseen input window.

\begin{figure*}[!h]
    \centering
    \includegraphics[width=\textwidth]{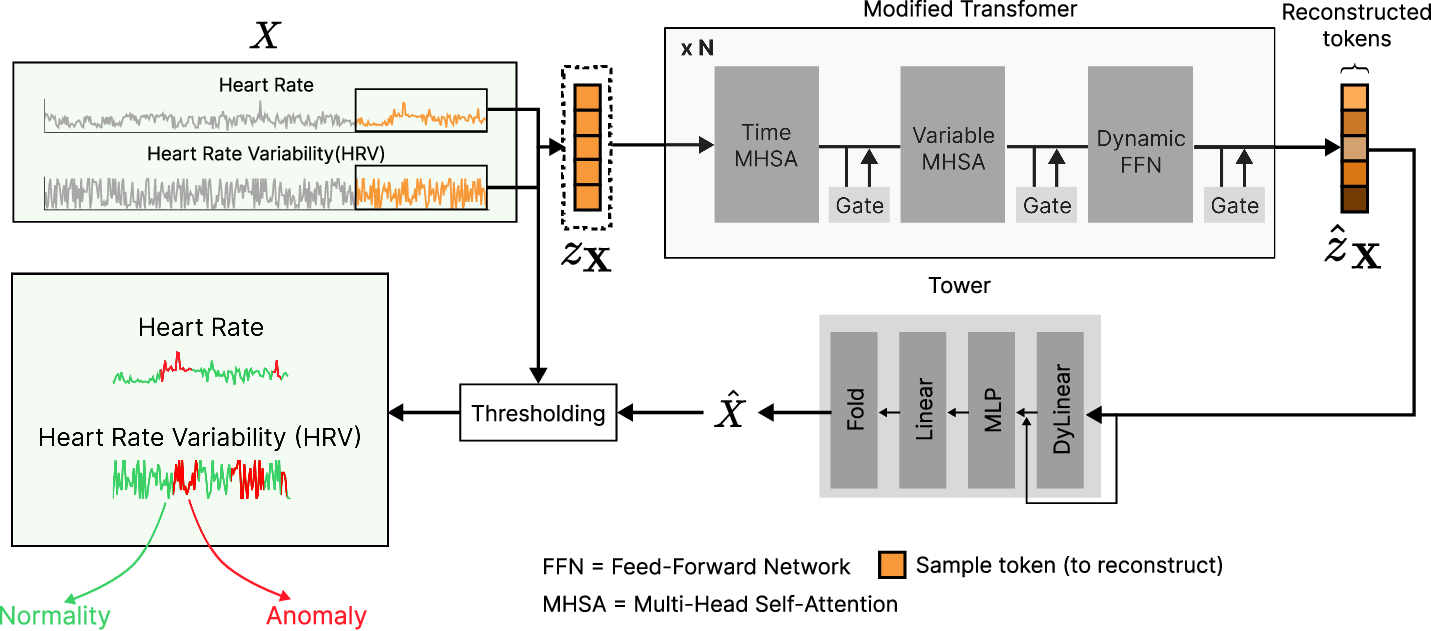}
    \caption{Fine-tuning of the UniTS architecture for anomaly detection. The multivariate time series in input gets transformed into several tokens, which are then passed through $N$ different blocks of the UniTS model. UniTS reconstructs the sample tokens, which are unpatched and compared against the real tokens. The comparison uses a dynamic threshold to discern between anomalies and normality. For visualization purposes, we do not illustrate the prompt tokens concatenated to the sample ones.}
    \label{fig:units}
\end{figure*}

\subsection{Framework Overview}\label{sub:framework}

UniTS -- see~\autoref{fig:units} -- is a versatile multi-task model featuring a unified network architecture.  Without loss of generality, we illustrate the fine-tuning process of  UniTS for anomaly detection, specifically using multivariate time series data like heart rate (HR) and heart rate variability (HRV). The process begins with the input time series data, $\mathbf{X}$, which contains both HR and HRV signals. This data is split into smaller segments or "tokens" passed to the UniTS model. The model consists of $N$ stacked blocks, each incorporating a modified transformer architecture. These blocks include key components like Time Multi-Head Self-Attention (Time MHSA), Variable MHSA, and Dynamic Feed-Forward Networks (Dynamic FFN). Gates control the flow of information between these components. Once processed by these blocks, the UniTS model outputs reconstructed tokens, represented as $\tilde{z}_\mathbf{X}$, which are then unfolded or unpatched into time series data. The reconstruction is compared to the original tokens using a dynamic threshold mechanism. This comparison helps identify anomalies in the time series. Normal HR and HRV signals are classified as ``normality,'' whereas significant deviations from the expected patterns are flagged as ``anomalies,'' as shown in the green and red segments. The dynamic threshold process ultimately discerns between normal and abnormal signals, visualizing anomalies for detection.

\paragraph{Input Tokenization} 
In UniTS, input tokenization is a foundational process that converts raw time series data into structured representations, enabling the model to process complex multivariate time series tasks effectively. Each token corresponds to either a temporal step or a variable, encapsulating temporal dynamics and inter-variable relationships. This tokenization scheme allows the model to apply attention mechanisms across time and variables, thereby improving its capacity to capture dependencies. The model avoids task-specific preprocessing by unifying inputs into tokens, ensuring generalization across forecasting, classification, imputation, and anomaly detection. The structured nature of tokenized inputs enhances adaptability, making it integral for multi-task learning. In more detail, UniTS introduces three\footnote{Although we do not use the task token in our scenario, we illustrate how they can be integrated into the framework. We point the reader to~\cite{gao2024units} for more details.} unique token types: i.e., \textit{sample},  \textit{prompt}, and \textit{task} tokens, each fulfilling a specific role in time series analysis.

\begin{itemize}
    \item \textit{Sample tokens.}  The time series\footnote{For simplicity, we assume that the entire time series $\mathbf{X}$ has already been split into multiple windows $\mathbf{W}$. Therefore, at inference, one reconstructs a single window at a time.} $\mathbf{X} \in \mathbb{R}^{T\times n}$.  We split $\mathbf{X}$ into patches along the time dimension via a non-overlapping patch size of $k$. Then, a linear layer projects each patch into an embedding vector of length $d$, obtaining sample tokens $z_{\mathbf{X}} \in \mathbb{R}^{\frac{T}{k} \times n \times d}$. $z_{\mathbf{X}}$ are added with learnable positional embeddings.

    \item \textit{Prompt tokens.} They are defined as learnable embeddings $z_p \in \mathbb{R}^{p \times n \times d}$ where $p$ is the number of tokens. These tokens incorporate the task UniTS needs to perform.  Notice that, in \autoref{fig:units}, we do not illustrate the prompt tokens proposed in the original paper since we only exploit the \textit{forecasting prompt} in our anomaly detection scenario. Nevertheless, for completeness, we invite the reader to visualize that these prompt tokens are prepended to the input series tokens as in Eq.~\eqref{eq:input_to_units}.
\begin{equation}\label{eq:input_to_units}
    z_\text{Anomaly} = [z_p, z_{\mathbf{X}}]_\mathcal{T} \in \mathbb{R}^{(p + \frac{T}{k}) \times n \times d},
\end{equation}
where $[*,*]_\mathcal{T}$ is the concatenation along the time axis. 

    \item \textit{Task tokens.} The UniTS framework comes with an additional type of tokens, namely the task ones. Since we want to reconstruct tokens to be able to detect anomalies, we do not explicitly rely on the task tokens as suggested in~\cite{gao2024units}. In this scenario, we want to compare the reconstructed tokens with those coming from the ground truth (GT) and pinpoint the anomalies along the series. In detail, instead of injecting task tokens into the input, we want to reconstruct the input tokens.
\end{itemize}

\paragraph{UniTS modules} The model takes in input $z_\text{Anomaly}$ and feeds it to $N$ blocks of modified transformer architecture to handle heterogeneous multi-domain data with varying dynamics and the number of variables. We use the original UniTS architecture, as shown in \autoref{fig:units}. In more detail, we use Time and Variable Multi-Head Self-Attention (MHSA) blocks, a Dynamic Feed-Forward Network (FFN), and gating modules. We refer the reader to the original paper's appendix~\cite{gao2024units} for more detail on the transformer and tower modules.

\begin{itemize}
    \item \textit{Time and Variable MHSA.} As suggested in the original paper, we use a two-way self-attention for feature and time dimensions. This approach contrasts with previous methods that apply self-attention to time or variable dimensions but not tensors. Time and variable self-attention effectively handle time series samples with various numbers of features $n$ and different time lengths $t$.

    \item \textit{Dynamic FFN.} The transformer block is modified by incorporating a dynamic operator into an FFN layer. This modification enables the FFN to capture dependencies between tokens, unlike the standard FFN, which processes embedding vectors point-wise.

    \item \textit{Gating.} To mitigate interference in the latent representation space, gating modules are used after each layer. This module dynamically re-scales features in layer-wise latent spaces and promotes the stability of latent representations.

    \item \textit{Tower.} This module transforms tokens into time points prediction results. Taking into consideration the generated sample tokens $\hat{z}_\mathbf{{X}}$, the tower module reconstructs the full time-series sample as in Eq.~\eqref{eq:rec_time_series}.
\begin{equation}\label{eq:rec_time_series}
    \hat{\mathbf{X}} = \text{Proj}(\text{MLP}(\hat{z}_\mathbf{{X}} + \text{DyLinear}(\hat{z}_\mathbf{{X}})))
\end{equation}
\begin{equation}
\begin{gathered}
    \text{DyLinear}(z_t, w) = \mathbf{W}_{\text{Interp}} z_t \in \mathbb{R}^{l_{\text{out}} \times d}\\
    \mathbf{W}_{\text{Interp}} = \text{Interp}(w) \in \mathbb{R}^{l_\text{in} \times l_\text{out}},
\end{gathered}
\end{equation}
where the MLP is composed of two linear layers with an activation layer in between,  Proj is
the unpatchify operation that transfers the embedding back to the time series patch; Interp is a bilinear interpolation to resize $w$ from shape $w_\text{in} \times w_\text{out}$ to $l_\text{in} \times l_\text{out}$.
\end{itemize}

\subsection{Fine-tuning UniTS for Abnormal Emotion Detection}
\label{sub:Fine-tuning}

UniTS was trained on a large, diverse dataset using a unified masked reconstruction scheme. This allows the model to learn general representations of temporal sequences and their underlying patterns, enhancing its generative and predictive task capabilities. Recall that UniTS is a foundational model in time series, and its training is executed on multiple tasks. In UniTS, whole model \textit{fine-tuning} and \textit{prompt-learning} represent two distinct paradigms for adapting the foundational model to new datasets and specific tasks, each with unique advantages and operational mechanisms.

\paragraph{Fine-tuning vs. Prompt-learning}

Fine-tuning involves updating the entire UniTS model parameters using the target dataset. This approach leverages the pre-trained representations learned during the initial training phase and adapts them to the nuances of the new data. Specifically, fine-tuning adjusts the weights of both the sequence and variable attention mechanisms and the dynamic linear operators within the unified network backbone. This comprehensive adaptation allows the model to specialize its understanding and prediction capabilities to better fit the characteristics of the new dataset. Prompt learning, on the other hand, maintains the integrity of the pre-trained UniTS model by keeping its weights frozen and introducing a set of learnable prompt tokens. These prompt tokens act as task-specific adapters that guide the model's processing without altering its foundational parameters. Only the embeddings of these prompt tokens are updated during prompt learning based on the target task, effectively steering the model to generate appropriate outputs for the new dataset. \autoref{tab:fine_tune_vs_prompt_learning} illustrates these paradigms' (dis)advantages.

\begin{table}[!h]
\centering
\caption{Comparison of Fine-tuning and Prompt-Learning in UniTS.}
\label{tab:fine_tune_vs_prompt_learning}
\resizebox{\textwidth}{!}{%
\begin{tabular}{@{}lll@{}}
\toprule &
  Fine-tuning &
  Prompt-Learning \\ \midrule
Advantages &
  \begin{tabular}[t]{@{}l@{}}\textcolor{mygreen}{\cmark} \,Full task-specific optimization. \\ \textcolor{mygreen}{\cmark} \,Enhanced performance for diverse datasets.\end{tabular} &
  \begin{tabular}[t]{@{}l@{}}\textcolor{mygreen}{\cmark} \, Efficient with frozen model weights. \\ \textcolor{mygreen}{\cmark} \, Rapid adaptation to new tasks.\\ \textcolor{mygreen}{\cmark} \,  Reduced risk of overfitting.\end{tabular} \\ \midrule
Limitations &
  \begin{tabular}[t]{@{}l@{}}\textcolor{myred}{\xmark} \, Expensive to update model parameters.  \\ \textcolor{myred}{\xmark} \, Overfitting on small datasets.\end{tabular} &
  \begin{tabular}[t]{@{}l@{}}\textcolor{myred}{\xmark} \, Lower performance vs. fine-tuning.\\ \textcolor{myred}{\xmark} \, Effectiveness depends on prompt design.\end{tabular} \\ \bottomrule
\end{tabular}%
}
\end{table}

According to Gao et al.~\cite{gao2024units}, prompt learning in UniTS achieves performance comparable to fine-tuning while incurring lower computational costs. This parity is particularly advantageous in applications where resource constraints are a concern or when deploying the model across multiple tasks simultaneously. However, fine-tuning remains the preferred approach when maximum performance on a specific task is paramount and computational resources are available.

In the context of our study, where the objective is to utilize HR and HRV signals for abnormal emotion detection via non-invasive wearable devices, the choice between fine-tuning and prompt learning is influenced by practical considerations. By preprocessing the datasets to obtain features at a low-frequency rate (e.g., transforming ECG samples from 700Hz to HR/HRV at 0.1Hz), we significantly reduce the computational burden of fine-tuning UniTS. This reduction aligns with our goal to emulate real-world scenarios where vital signals are measured with smartwatches, balancing effective battery usage with reliable signal acquisition~\cite{10.1145/2674396.2674446}. Consequently, fine-tuning becomes a more feasible and  ``affordable'' strategy in our application, enabling the model to adapt effectively without compromising efficiency. We follow~\cite{10.1145/2674396.2674446} to reduce the sampling rate -- this is also supported in~\autoref{fig:sampling_intervals} of \autoref{qualitative}. Finally, we use 20 epochs and an exponential learning rate decay to finetune the model.

\subsection{Adding explainability features}
\label{subsec:explainabilityLLM}

Once the UniTS model is finetuned for anomaly detection, we leverage the output of the Tower module during inference (as illustrated in~\autoref{fig:units}) to serve as input for a Large Language Model (LLM), such as GPT-4o~\cite{achiam2023gpt} or LLaMa-2~\cite{touvron2023llama}, to generate explanations for the detected anomalies. Specifically, the anomaly score generated by UniTS represents the likelihood of an anomaly, calculated as the ratio of detected anomaly energy to a set threshold. This score indicates the intensity of the anomaly, with values below 1 denoting normal samples and values above 1 indicating an anomaly. This score is then used to prompt the LLM to generate interpretive explanations for domain experts.

The LLM can be prompted to contextualize its explanations according to the audience, ensuring the generated output is tailored to the user's expertise level. For instance, as noted by~\cite{DBLP:journals/corr/abs-2402-18180}, persona-based interaction allows the LLM to adjust the technicality of the explanation based on the user's role, such as a clinician or a system administrator. The prompt informs the LLM in the example that it is ``interacting'' with a professional clinician, thereby guiding it to generate explanations using clinical terminology and domain-specific knowledge. This capability is critical in clinical settings, where precision and appropriate medical jargon are necessary to convey the physiological implications of the detected anomalies. Moreover, LLM-based explanations can support multi-level decision-making by presenting the expected physiological norms alongside detected anomalies. This dual approach ensures that stakeholders are alerted to deviations and informed of the baseline expectations, facilitating a deeper understanding of abnormal and normal readings. The illustrative example in~\autoref{fig:prompt} demonstrates this by asking the LLM to explain why the predicted anomalies occur while requesting the expected physiological behavior when anomaly scores exceed a specified threshold. Here, the system is calibrated to interpret HR and HRV signals, providing reasoning for abnormal readings and expected ``normal'' physiological patterns using medical language appropriate for the clinical domain.

\section{Experiments and Results}

\subsection{Experimental Setup}\label{sec:exp_setup}

\paragraph{Compared methods}
Notice that since the works in emotion monitoring do not model stress detection as an anomaly detection problem (see~\autoref{tab:sota}), we compare only with the works surveyed in~\autoref{sec:ad_in_time_series}. We compare supervised UniTS and fine-tuned UniTS with various SoTA models for multivariate time-series anomaly detection. The models compared include TranAD~\cite{tuli2022tranad}, LSTM-NDT~\cite{hundman2018detecting}, CAE-M~\cite{zhang2021unsupervised}, DAGMM~\cite{zong2018deep}, USAD~\cite{audibert2020usad}, OmniAnomaly~\cite{su2019robust}, MTAD-GAT~\cite{zhao2020multivariate}, GDN~\cite{deng2021graph}, MAD-GAN~\cite{Li2019MADGANMA}, MSCRED~\cite{zhang2019deep}, HypAD~\cite{flaborea2023we}, and TadGAN~\cite{geiger2020tadgan}. We adopted the hyperparameters presented in their original papers for these models.

\paragraph{Training UniTS} For fine-tuning the UniTS model, we utilized a window size of 5, an embedding dimension of 128, and a learning rate of $5 \times 10^{-4}$. The datasets were split into 80\% training and 20\% testing sets. All models were trained for 20 epochs, and the F1 score was calculated as the average across a 5-fold cross-validation.

\paragraph{Fair comparisons and threshold settings} We systematically preprocessed our datasets to ensure consistency and comparability across different sources. From all datasets, we extracted HR and HRV every 10 seconds using a sliding window approach based on the preceding 60 seconds of signal data from ECG and, when available, Blood Volume Pulse (BVP), which measures the

\begin{figure}[H]
    \begin{boxA}
        \scriptsize
        \textcolor{black}{
            \textbf{Prompt:}\newline
            You are an automated system that interprets changes in HR (Heart Rate) and HRV (Heart Rate Variability) signals when emotional anomalies are detected in a patient's vital signs with respect to a resting state condition. \newline
            The data provided includes:
            \vspace{-0.8em}
            \begin{itemize}
                \setlength{\itemsep}{-3pt}
                \item \textbf{Time}: Seconds since the start of the recording
                \item \textbf{HR}: Heart Rate in beats per minute
                \item \textbf{HRV}: Heart Rate Variability (RMSSD) in milliseconds
                \item \textbf{Prediction}: Indicates if an anomaly was detected (1) or not (0)
                \item \textbf{Anomaly Energy}: Ratio of the detected anomaly energy to the threshold, indicating the intensity of the anomaly. If energy is $ < 1$, the sample is normal; otherwise, it indicates an anomaly.
            \end{itemize}
            \vspace{-0.8em}
            The patient is a young adult with no known medical conditions. The data was collected using a wearable ECG sensor.\newline 
            \vspace{0.5em}Patient Data: \newline
            \begin{tabular}{c c c c c}
                \textit{Time} & \textit{HR} & \textit{HRV} & \textit{Prediction} & \textit{Anomaly Energy} \\
                0 & 82.39 & 20.9 & 0 & 0.0 \\
                \vdots & \vdots & \vdots & \vdots & \vdots \\
                80 & 82.56 & 10.8 & 1 & 1.3 \\
            \end{tabular}\newline \newline
            Provide a concise explanation structured in four sections: \textbf{\colorbox{mygreen}{\textcolor{white}{Analysis}}}, \textbf{\colorbox{myblue}{\textcolor{white}{Causes}}}, \textbf{\colorbox{myred}{\textcolor{white}{Criticality}}}, and \textbf{\colorbox{myorange}{\textcolor{white}{Scores}}}.
            \vspace{-0.8em}
            \begin{itemize}
                \setlength{\itemsep}{-3pt}
                \item \textbf{\colorbox{mygreen}{\textcolor{white}{Analysis}}}: Highlight any strong, high-intensity anomalies if present, using a dotted list to describe the observed changes in HR and HRV signals. Avoid directly mentioning energy values; instead, describe the anomalies’ clinical relevance and intensity. Describe any trends in HR and HRV over time, noting persistent patterns or shifts in the data that may indicate physiological or emotional responses.
                \item \textbf{\colorbox{myblue}{\textcolor{white}{Causes}}}: Propose likely causes of these anomalies, considering the patient's demographic, the characteristics of the detected anomalies (e.g., short, sudden, prolonged), and any environmental or measurement factors that could affect signal accuracy. Distinguish between potential true physiological changes, system errors, or data noise.
                \item \textbf{\colorbox{myred}{\textcolor{white}{Criticality}}}: Assess the clinical significance of these anomalies. Consider both the impact of individual anomalies (especially high-energy ones) and the cumulative effect if anomalies are frequent or prolonged. Specify if the anomaly could require immediate follow-up, if it's non-critical but worth monitoring, or if it seems likely to be insignificant or an error.
                \item \textbf{\colorbox{myorange}{\textcolor{white}{Scores}}}:
                    \vspace{-0.8em}
                    \begin{itemize}
                        \setlength{\itemsep}{-3pt}
                        \item \textbf{Significance Score} (0-10): Rate the significance of the anomaly based on how clinically relevant it is, with 0 being not relevant (e.g., noise) and 10 being highly relevant (e.g., real physiological change).
                        \item \textbf{Criticality Score} (0-10): Rate the potential clinical impact or urgency, with 0 being minimal concern and 10 being critical. This should consider both the intensity of the anomaly and the risk it may pose to the patient's well-being.
                    \end{itemize}
            \end{itemize}
            \vspace{-0.8em}
            Ensure the explanation is brief, clinically relevant, and suitable for interpretation by a professional clinician.
        }
    \end{boxA}
    \caption{Prompt used to interpret HR and HRV signals, focusing on emotional anomalies, and generate a structured clinical analysis for patient data interpretation.}
\label{fig:prompt}
\end{figure}
\noindent volumetric change in blood circulation through the microvascular bed of tissue. BVP signals can be extracted non-invasively using photoplethysmography (PPG), a technique commonly integrated into smartwatches and other wearable devices. This capability is instrumental in stress detection and management, making wearable devices valuable in clinical and everyday settings. Lastly, we use a threshold of the 3rd and 97th percentile to indicate anomalous data points: i.e., everything below the 3rd or above the 97th percentile is labeled as 1; otherwise, it is 0. Percentiles are calculated on the reconstruction errors.

\subsection{Datasets}\label{sec:datasets}
We use the following three widely adopted datasets for emotion detection:
\begin{itemize}
    \item \textit{DREAMER}~\cite{katsigiannis2017dreamer} is designed for emotion recognition via low-cost devices. It consists of EEG and ECG readings recorded from 23 participants. ECG signals were captured at a sampling rate of 256 Hz using a SHIMMER wireless sensor. Movie clips were used to stimulate different emotions in participants, and at the end of each clip, participants scored their self-assessment on a 5-point arousal, valence, and dominance scale. Here, we treat states with a valence of 1 as abnormal and the rest as normal.

    \item \textit{MAHNOB-HCI}~\cite{soleymani2011multimodal} was collected using the Biosemi Active II system with active electrodes to capture physiological signals, including ECG and EEG, from 27 participants at a sampling rate of 256 Hz. Participants were asked to record their emotional status after each trial, consisting of multimedia content stimulation. Here, we treat the states with a valence of 1 as abnormal and the rest of the emotional states as normal. For our experiments, we selected subsets of the dataset with recordings of at least 120 seconds and excluded those with incomplete tracks due to recording errors.
    
    \item \textit{WESAD}~\cite{schmidt2018introducing} has ECG and BVP signals from 15 participants at a rate of 700Hz extracted via a RespiBAN Professional sensor worn on the chest and the Empatica E4 wristband. It contains four different emotional states. Here, we treat the stress state as abnormal and the rest as normal. The stress condition is obtained by exposing the subjects to the Tier Social Stress Test, which consists of public speaking and a mental arithmetic task.
\end{itemize}
Note that all datasets induce emotions in a controlled setting, a limitation that we discuss in~\autoref{qualitative}.

\begin{table}[!t]
\centering
\caption{Comparison of different HR and HRV methods extracted from ECG and BVP. Bold-faced values illustrate the best; underlined ones are the second best.}
\label{tab:comparison}
\resizebox{\linewidth}{!}{%
\begin{tabular}{@{}lcccc|c@{}}
\toprule
 
{} &
  {DREAMER} &
  {HCI} &
  {\begin{tabular}[c]{@{}c@{}}WESAD\\ (ECG)\end{tabular}} &
  {\begin{tabular}[c]{@{}c@{}}WESAD\\ (BVP)\end{tabular}} &
  {\begin{tabular}[c]{@{}c@{}}AVG\\ F1\end{tabular}} \\ \midrule
LSTM-NDT~\cite{hundman2018detecting}    & 0.235 & 0.373 & 0.694 & 0.760 & 0.515 ± 0.218 \\
MSCRED~\cite{zhang2019deep}      & 0.434 & 0.618 & 0.594 & 0.536 & 0.545 ± 0.071 \\
HypAD~\cite{flaborea2023we}       & 0.513 & 0.547 & 0.669 & 0.479 & 0.552 ± 0.072 \\
TranAD~\cite{tuli2022tranad}      & 0.515 & 0.622 & 0.474 & 0.717 & 0.582 ± 0.095 \\
OmniAnomaly~\cite{su2019robust} & 0.543 & 0.513 & 0.716 & 0.583 & 0.589 ± 0.078 \\
MTAD-GAT~\cite{zhao2020multivariate}    & 0.369 & 0.677 & 0.757 & 0.676 & 0.620 ± 0.149 \\
CAE-M~\cite{zhang2021unsupervised}      & 0.634 & 0.513 & \underline{0.762} & 0.590 & 0.625 ± 0.090 \\
MAD-GAN~\cite{Li2019MADGANMA}     & 0.390 & \underline{0.678} & 0.735 & 0.750 & 0.638 ± 0.146 \\
TadGAN~\cite{geiger2020tadgan}      & 0.528 & 0.657 & 0.705 & 0.680 & 0.643 ± 0.068 \\
USAD~\cite{audibert2020usad}        & 0.590 & 0.641 & 0.686 & 0.759 & 0.669 ± 0.062 \\
GDN~\cite{deng2021graph}         & \underline{0.712} & 0.501 & 0.697 & \underline{0.778} & 0.672 ± 0.103 \\
DAGMM~\cite{zong2018deep}       & 0.619 & 0.596 & 0.744 & 0.733 & \underline{0.673 ± 0.066} \\
\midrule
UniTS~\cite{gao2024units}       & \textbf{0.828} & \textbf{0.864} & \textbf{0.793} & \textbf{0.800} & \textbf{0.821 ± 0.049} \\
\bottomrule
\end{tabular}%
}
\end{table}

\begin{figure}[!t]
    \begin{minipage}[t]{0.33\textwidth}
    \flushleft
        \captionof{table}{P-values produced by the Dunn post-hoc test, where the control model is UniTS against the top-performing SoTA methods. We use a p-value of $0.01$ to reject the null hypothesis.}
        \label{tab:dunn_p_values}
        \resizebox{\textwidth}{!}{%
        \begin{tabular}{@{}lc@{}}
        \toprule
                 & UniTS                            \\ \midrule
        DAGMM    & $1.300 \times 10^{-3}$ \\ 
        GDN      & $1.995 \times 10^{-3}$        \\
        USAD     & $9.830 \times 10^{-4}$  \\
        TadGAN   & $1.940 \times 10^{-4}$   \\
        MAD-GAN  & $7.760 \times 10^{-4}$  \\
        \bottomrule
        \end{tabular}%
        }
    \end{minipage}%
    \hfill
    \begin{minipage}[t]{0.63\textwidth}
        \adjustbox{valign=t}{\includegraphics[width=\textwidth]{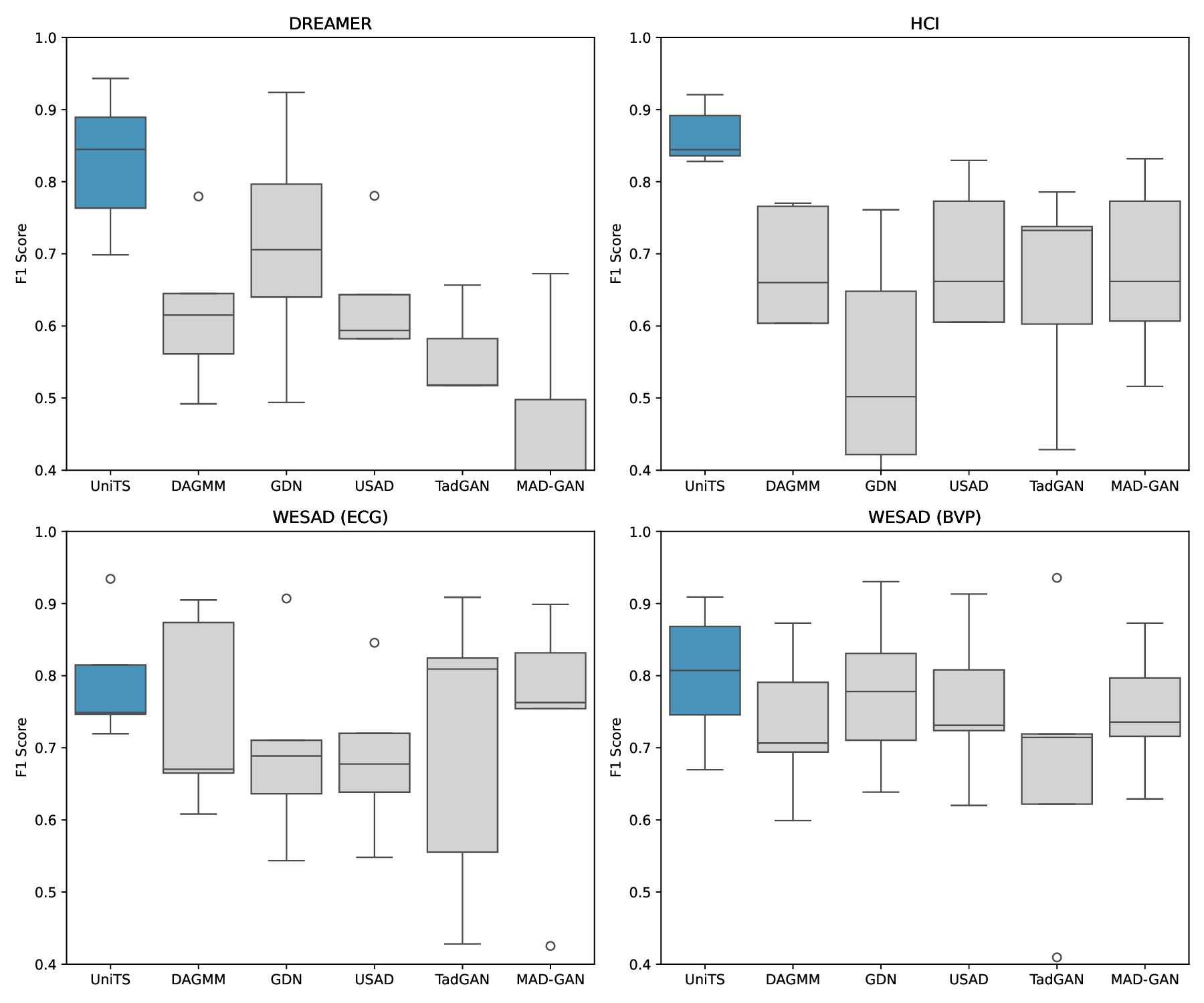}}
        \captionof{figure}{Confidence intervals for the performances of UniTS and the best SoTA methods as per the Dunn post-hoc test. We highlight the method that performs best on average according to the values reported in~\autoref{tab:comparison}.}
        \label{fig:boxplot}
    \end{minipage}
\end{figure}

\begin{figure}[!t]
    \centering
    \begin{subfigure}{0.45\textwidth}
    \includegraphics[width=\textwidth]{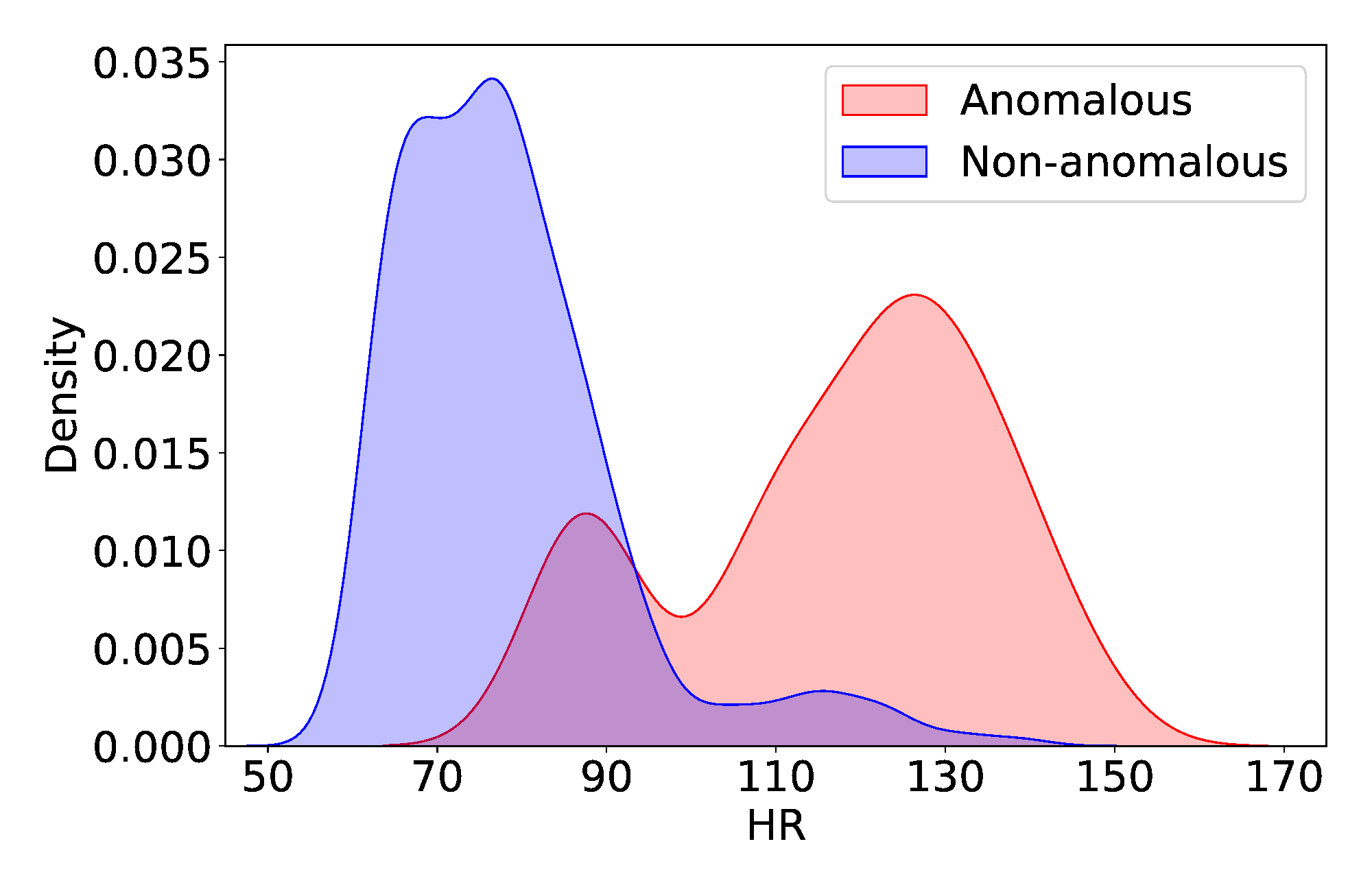}
    \end{subfigure}%
    \begin{subfigure}{0.45\textwidth}
        \includegraphics[width=\textwidth]{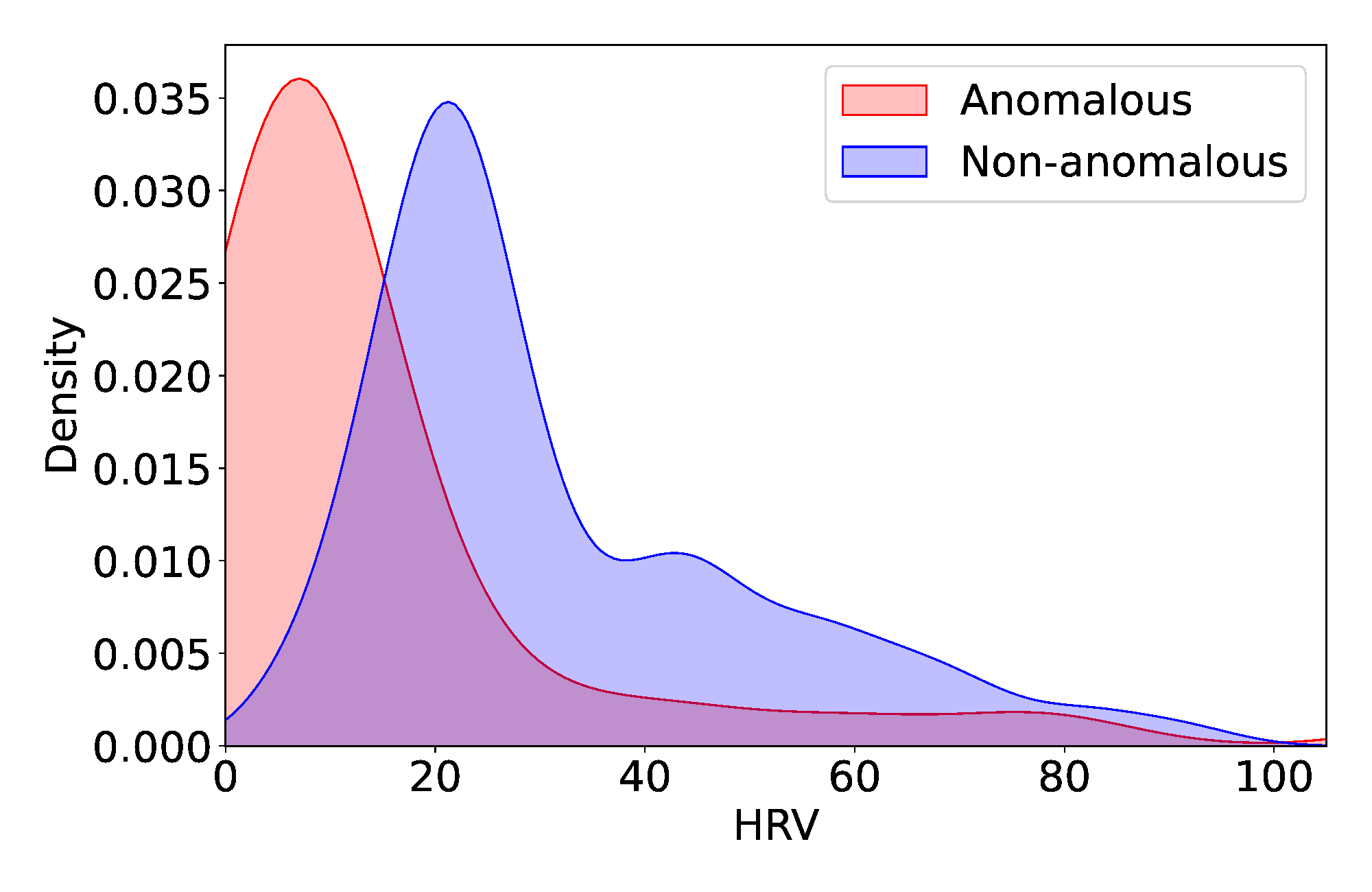}
    \end{subfigure}
    \begin{subfigure}{0.45\textwidth}
        \includegraphics[width=\textwidth]{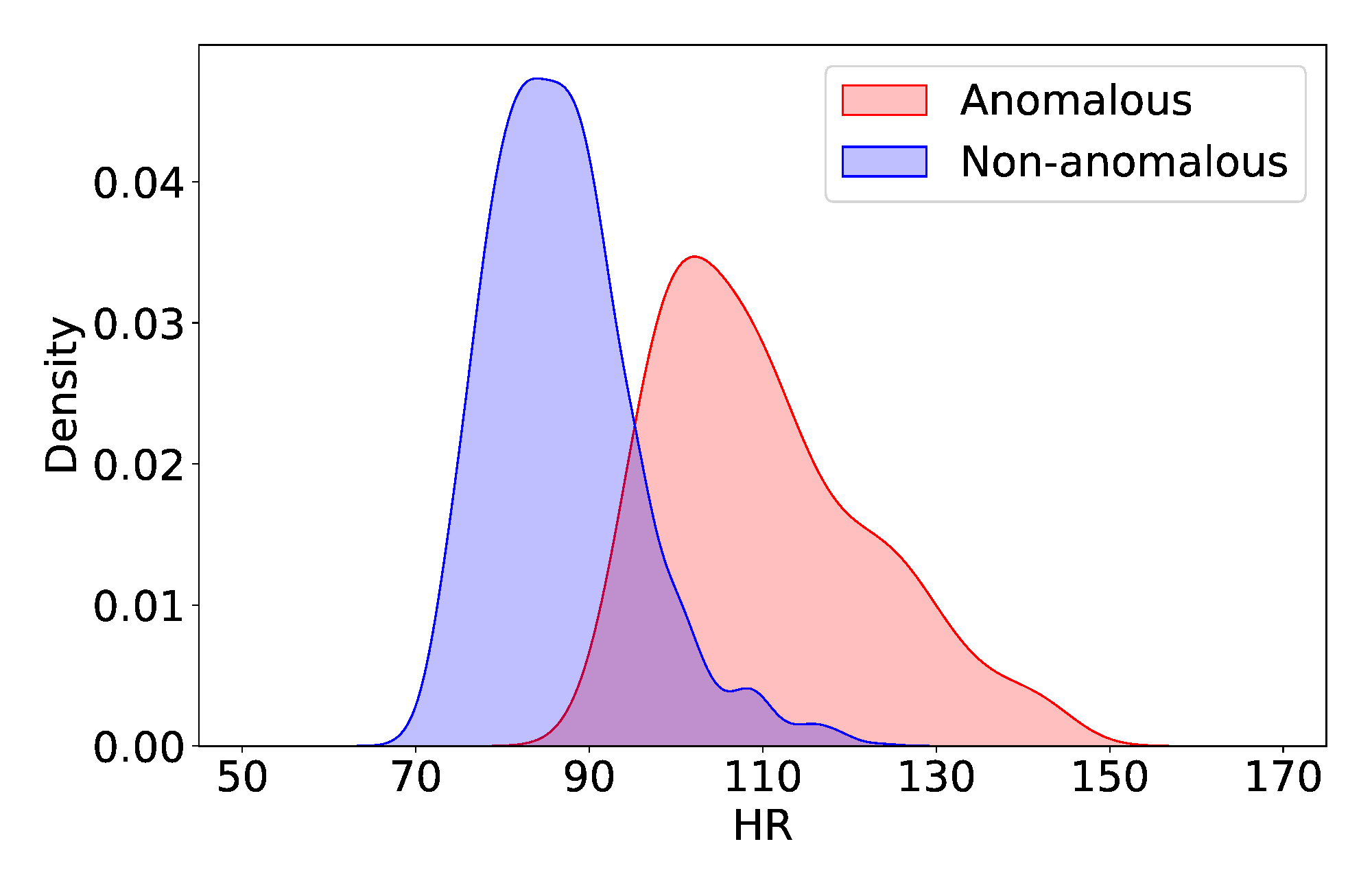}
    \end{subfigure}%
    \begin{subfigure}{0.45\textwidth}
        \includegraphics[width=\textwidth]{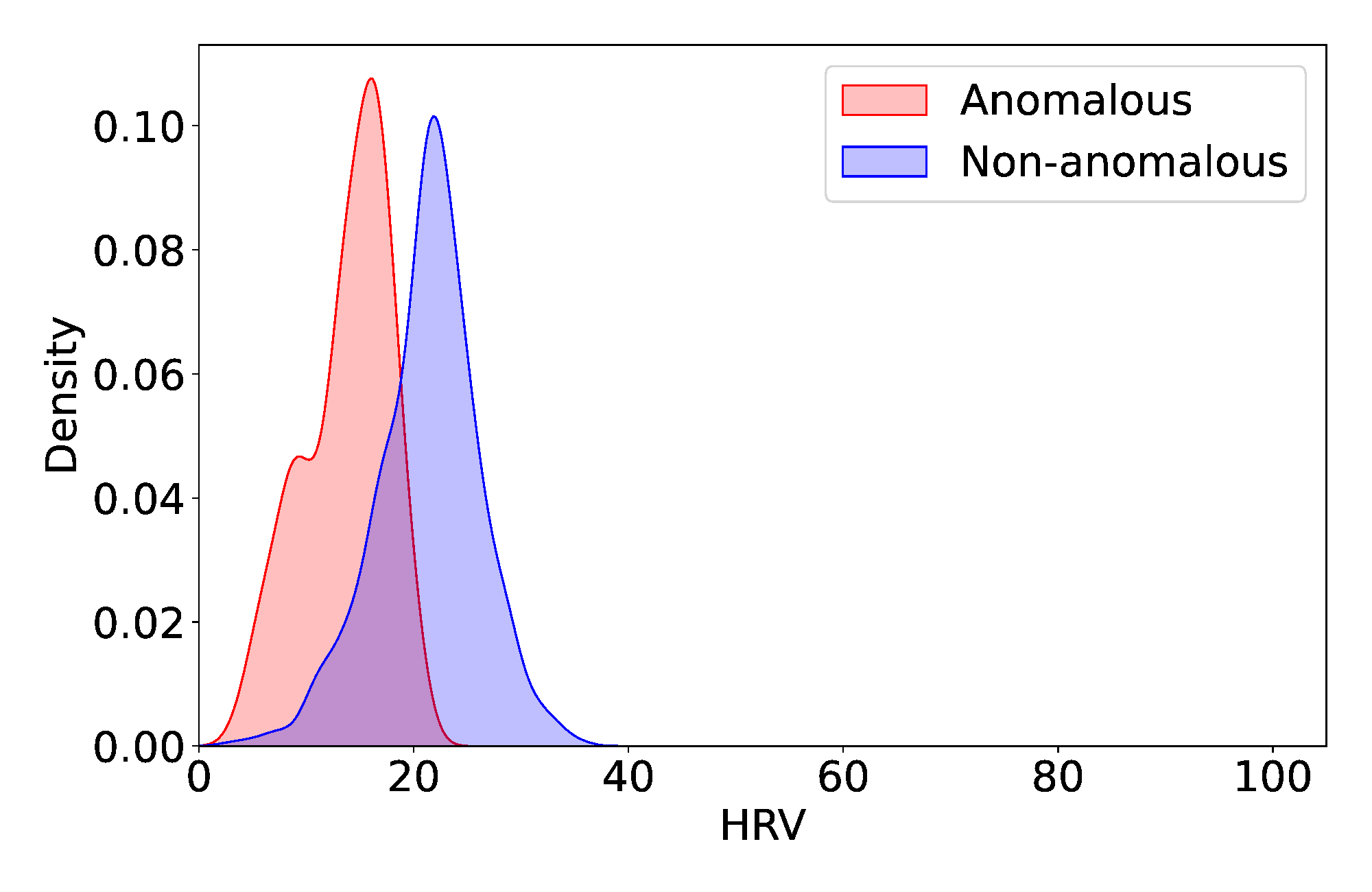}
    \end{subfigure}
    \caption{Density plots for HR and HRV on the test set for WESAD for ECG (up) and BVP (down) versions. We also show the means of the distributions with dashed lines.}
    \label{fig:density}
\end{figure}

\subsection{Quantitative  Results}
\label{quantitative}

\paragraph{UniTS surpasses SoTA anomaly detection systems and shows remarkable stability and robustness across datasets} As shown in~\autoref{tab:comparison}, UniTS outperforms all the compared systems, achieving a considerable average improvement of $22\%$ on the second-best -- i.e., $16.29\%$ on DREAMER, $27.43\%$ on HCI, $4.07\%$ on WESAD (ECG), and $3.53\%$ on WESAD (BVP). We first conduct a Friedman Test on the top methods (UniTS, DAGMM, GDN, USAD, TadGAN, MAD-GAN) to show that UniTS has, statistically and significantly, the best performance across the board ($F=20.911$, $p=.0008 < .01$). Then, we perform a Dunn post-hoc test where the control anomaly detector is UniTS. The test suggests that UniTS is statistically and significantly different (better) across the board on average (see Table~\ref{tab:dunn_p_values}). 
We also note that the competitive advantage of UniTS is lower in the WESAD dataset. Several systems achieve more or less similar - and in some cases slightly better - performances. We argue that WESAD is the only dataset that explicitly includes stress among the classified emotions. In contrast, the other two datasets use dimensional models of emotions (arousal and valence), resulting in a nuanced and less ``clear-cut'' classification. In this simpler context, many systems perform much better.
BVP can be noisy and contain many artifacts, which hardens extracting meaningful features from it~\cite{10.1145/3027063.3053121}. Hence, in WESAD (BVP), the normal and abnormal data distribution is noised and does not respect that originally coming from the ECG -- i.e., compare HR and HRV distributions in~\autoref{fig:density} and notice the difference in scale in HRV coming from ECG and BVP -- which we argue leads SoTA methods to underperform. Furthermore, UniTS, through its incorporated attention mechanism over the variates and the time axes, mitigates this phenomenon by giving less weight to noisy samples, maintaining high WESAD (BVP) performances.\footnote{We verified that UniTS has an FPR of $0.031$ in both versions of WESAD. Meanwhile, it reports an FNR of $0.172$ in WESAD (ECG) and $0.159$ in WESAD (BVP), suggesting that it is more capable of detecting the true anomalies (i.e., it has better recall).} Overall,  UniTS exhibits remarkably stable performance and higher robustness across all datasets and systems (std of only $.049$), resulting from being a universal model pre-trained on many diverse signals and tasks. 

For completeness purposes, in~\autoref{fig:boxplot}, we provide the reader with confidence intervals on the performances of UniTS and five top-performing SoTA methods according to the Dunn post-hoc test analysis -- see Table \ref{tab:dunn_p_values}.

\paragraph{UniTS achieves performances from lightweight devices comparable to SoTA systems using more invasive and potentially accurate sensors} The only dataset collecting signals from ECG devices and wristbands (BVP) is WESAD. As shown in~\autoref{tab:comparison}, UniTS delivers consistently high F1 scores with both ECG and BVP and maintains similar results across both types, while some other systems experience varied performance with BVP signals. This is a valuable result since, as discussed in~\autoref{sec:introduction}, lightweight devices, such as wristbands and smartwatches, are less expensive and invasive, enabling seamless monitoring. Using non-invasive, wearable technology without sacrificing performance can expand stress-related research and applications. Future studies can integrate these devices in various settings to gather continuous stress data, leading to personalized stress management programs, early intervention systems for mental health, and better well-being monitoring.

\begin{figure}[!t]
    \centering
    \includegraphics[width=\linewidth]{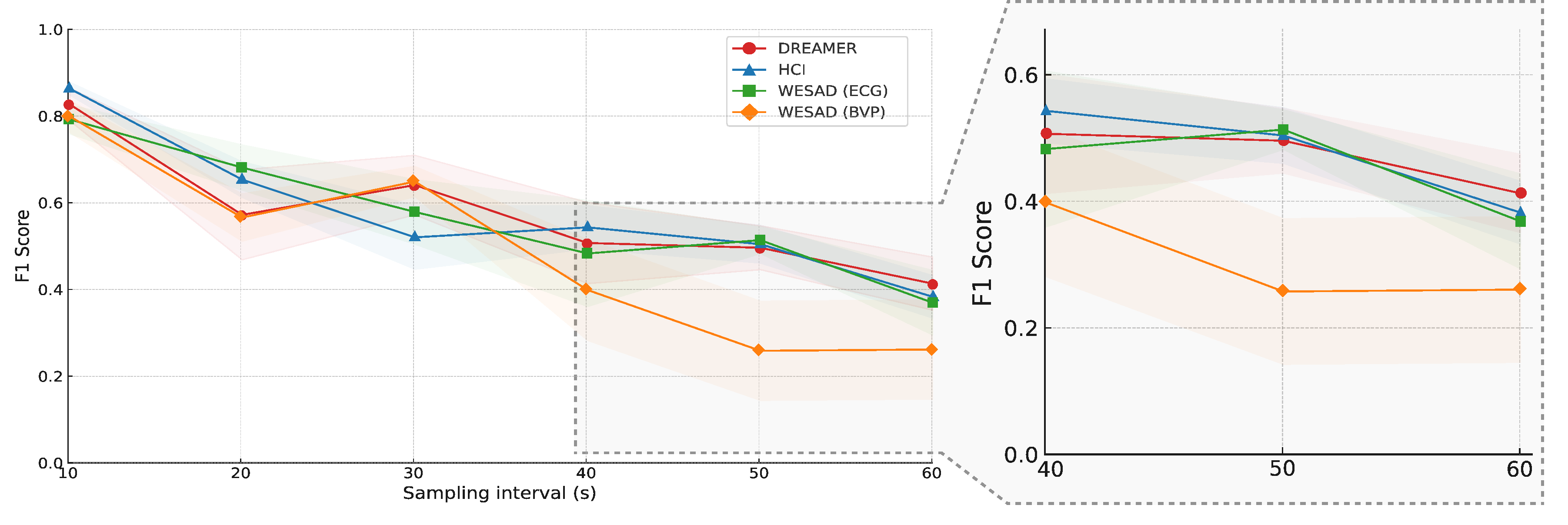}
    \caption{Performance degradation of F1 score across different sampling intervals using the UniTS on all the datasets. The zoomed-in area illustrates how the overall decreasing performance trend starts to have diminishing effects, especially in WESAD (BVP).}
    \label{fig:sampling_intervals}
\end{figure}

\paragraph{Lower sampling rates on lightweight devices make performance plummet by $\sim$$54\%$} \autoref{fig:sampling_intervals} shows the importance of the sampling rate on all datasets on ECG and BVP versions. Here, we illustrate the trend of average F1 scores of UniTS when the sampling interval increases from 10 to 60 seconds. The vital signals (i.e., HR and HRV) are sampled every $k$ seconds to build the multivariate time series taken in input, with varying $k$. As expected, the longer the sampling interval, the less responsive the model becomes, resulting in a degradation of results. This phenomenon is even more visible when we extract the HR and HRV from BVP, having a drop of $-53.89\%$ passing from a 10 to a 60-second sampling rate. This discrepancy highlights a critical consideration in the evolving landscape of wearable technology and continuous monitoring systems.
Therefore, we suggest re-evaluating the balance between device convenience and performance. For applications requiring high precision, such as medical diagnostics or performance monitoring in athletes, adhering to more frequent sampling intervals appears imperative. Therefore, our study advocates for a nuanced approach, where the selection of sampling intervals is tailored to the specific requirements of the use case, ensuring that the constraints of the monitoring device do not compromise the integrity of physiological data.

\subsection{Qualitative Results}
\label{qualitative}

\begin{figure}[!t]
    \centering
    \includegraphics[width=\linewidth]{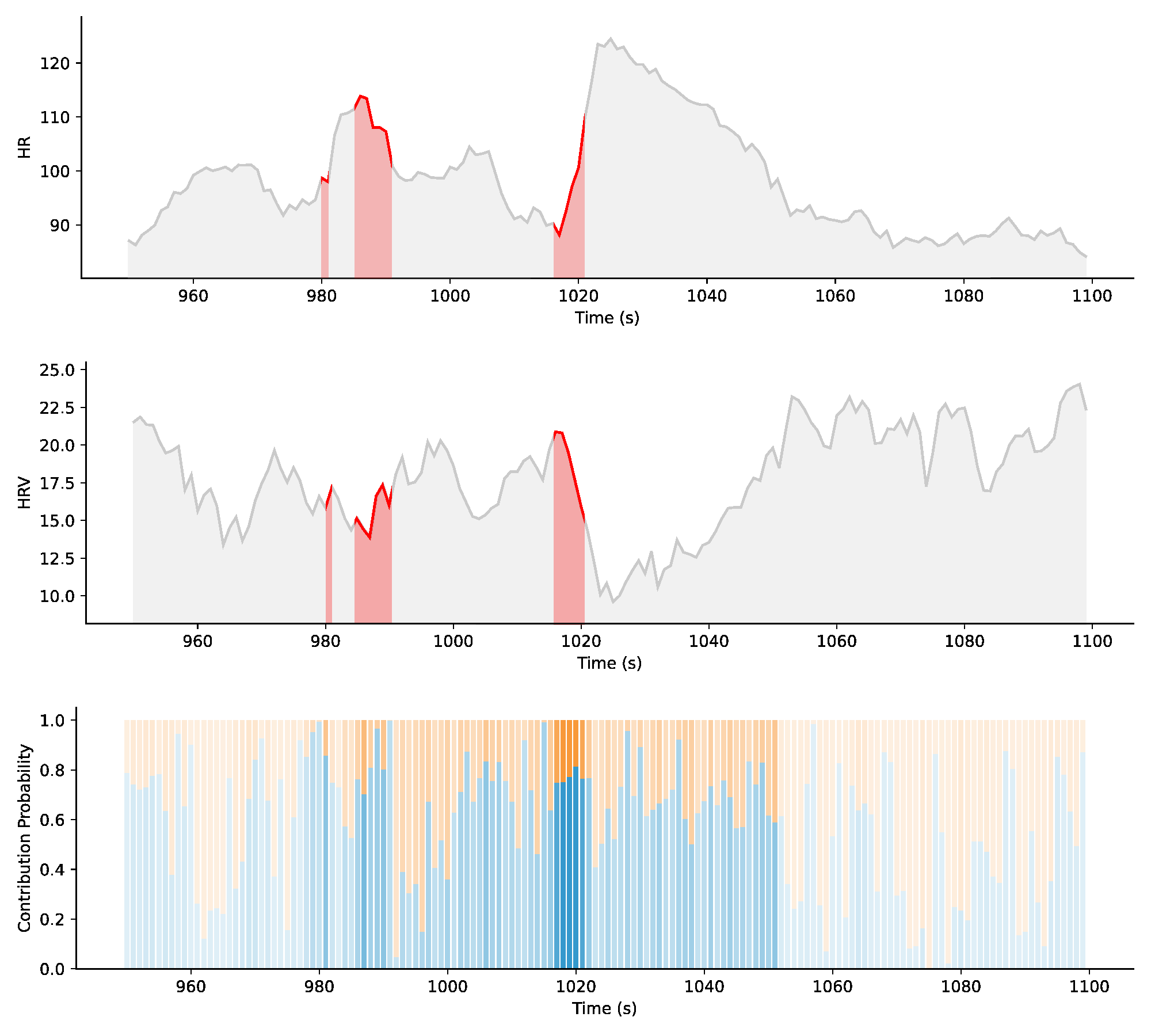}
    \caption{HR (up) and HRV (middle) contribution in WESAD (BVP) during inference time.  Red-shaded areas depict the reported anomalies. The lower bar plot illustrates the contribution probabilities based on the reconstruction error of each signal. The more transparent the contribution, the lower the chances of the signals being considered anomalous; the more opaque the contributions are, the more the chances of the signals being anomalous. Note that the transparency can be seen as visual uncertainty about the data point at a particular time step.}
    \label{fig:hr_vs_hrv}
\end{figure}

\paragraph{Casting the problem as one of anomaly detection enhances explainability}  Anomaly detection begins by establishing a model of ``normal'' behavior based on the monitored signals. Next, it detects deviations from this normality. This helps clinicians understand the baseline deviations identified and align better with medical professionals' needs and practical constraints. First, clinicians are usually more interested in identifying and investigating abnormalities than confirming the norm. Anomaly detection systems align with doctors' diagnostic workflow by focusing on detecting deviations from the norm. Secondly, these models provide significant benefits for precision medicine and personalization due to their ability to identify unique patient profiles and adapt to new data continuously~\cite{personalized}. Furthermore, in the case of multiple monitored signals, the model should highlight which ones are more indicative of stress~\cite{nature24}. In detail, although anomaly detectors are not interpretable per se,\footnote{Without loss of generality, anomaly detectors produce reconstruction errors for each data point, combining all feature dimensions. These errors do not necessarily indicate which features/signals contribute to the anomalousness of the data point itself.} UniTS reconstructs the multivariate time series to provide reconstruction errors for each series' dimension. This inherent characteristic allows us to interpret whether anomalies occurred, which is paramount, and which features contributed to their detection.

For example,~\autoref{fig:hr_vs_hrv} shows how to support doctors in identifying which signals contributed the most to detecting anomalies. Here, we depict the contribution probability based on the reconstruction error of each signal -- i.e., HR (up) and HRV (middle) -- in determining anomalies. The more opaque the shade of each bar in the plot, the more likely an anomaly will occur. Notice how the opacity of the probabilities increases when there is an abrupt trend change in one (or both) of the signals, which aligns with the red areas depicting the detection of an anomaly. One can think of the transparency of each bar as a visual uncertainty of the model. Therefore, the more opaque the bars are, the more certain the model is that there is an anomaly in that particular time frame. Meanwhile, when the bars are transparent, they may indicate two things: either the model is uncertain, or the data points therein are normal. To this end, we integrate the contribution probability with detecting anomalies in the raw signals. Additionally, we invite the reader to notice the different transparency levels within the time frame where an anomaly is detected. This phenomenon is expected as not all anomalies (gradual vs. abrupt) are detected with the same confidence. Interestingly, badly reconstructing HR in this scenario pushes the model to be more confident about an anomaly occurring -- see the middle portion of the third row in~\autoref{fig:hr_vs_hrv}. Contrarily, even if UniTS cannot effectively reconstruct HRV, it does not lead to detecting anomalies -- e.g., notice the more transparent tail-ends of the probabilities plot. By exploiting these types of explicit and inherently interpretable visualizations -- i.e., the contribution probability is estimated via the reconstruction error -- doctors can interpret which signal guides the prediction of an anomaly during inference and intervene accordingly. Finally, this type of analysis can be used to debug if the monitoring system functions correctly. For instance, if the contribution probability bars are opaque, yet the system does not report an anomaly, there might be something wrong with the current monitoring setup. In addition to a graphical representation of the anomaly (showing the dimensions that contribute most to the detected anomaly), we provide clinicians with a rich textual explanation, as detailed below.

\paragraph{Zero-shot integration with LLMs further enhances explainability}
\label{par:zeroshotllm}
In real-world scenarios, anomalies detected by the model require further investigation by a medical professional to ensure their correctness and assess whether they could result in any significant health risks for the patient. The anomaly detection model offers useful insights, but these findings require more context and interpretation. Using an LLM, we can provide detailed explanations that help clarify the anomaly and its potential impact, as illustrated in~\autoref{fig:dreamer-example} and~\autoref{fig:hci-example} of  \ref{app:qualitative}.
Here, we evaluated the utility and accuracy of explanations generated by GPT-4o using qualitative assessments from a medical professional. The explanations, derived from the outputs of the anomaly detection model, were structured into four sections: Analysis, Causes, Criticality, and Scores, according to the scheme illustrated in~\autoref{fig:prompt}. We generated an evaluation set of 16 explanations, which were reviewed by a highly experienced doctor,\footnote{With more than 40 years of experience.} who thoroughly examined the clinical data and images before considering the textual explanations provided by the system. Based on her evaluation, we assessed the explanations' relevance, coherence, and specificity, with the results summarized in~\autoref{tab:explaination-eval}.

\begin{table}[!t]
\centering
\caption{Qualitative evaluation by a medical professional of GPT-4o's explanations, assessing the usefulness of each analysis, causes, criticality, and the coherence and specificity of the overall explanation. Ratings are on a 0-5 scale. Each dataset contains 4 different explanations. Here, we show averages for each dataset and the cumulative average ($\mu \pm \sigma$) over all explanation evaluations.}
\label{tab:explaination-eval}
\resizebox{\linewidth}{!}{%
    \begin{tabular}{@{}lccc|cc@{}}
\toprule
 &
  \textbf{\begin{tabular}[c]{@{}c@{}}Utility of\\\colorbox{mygreen}{\textcolor{white}{Analysis}}\end{tabular}} &
  \textbf{\begin{tabular}[c]{@{}c@{}}Utility of\\ \colorbox{myblue}{\textcolor{white}{Causes}}\end{tabular}} &
  \textbf{\begin{tabular}[c]{@{}c@{}}Utility of \\ \colorbox{myred}{\textcolor{white}{Criticality}}\end{tabular}} &
  \textbf{\begin{tabular}[c]{@{}c@{}}Explanation\\ Coherence\end{tabular}} &
  \textbf{\begin{tabular}[c]{@{}c@{}}Explanation\\ Specificity\end{tabular}} \\ \midrule
  DREAMER & 5.00 & 4.75 & 4.50 & 4.25 & 1.25\\
  HCI & 5.00 & 4.75 & 4.50 & 4.50 & 0.50 \\
  WESAD (ECG) & 4.75 & 4.25 & 4.50 & 4.50 & 0.25\\
  WESAD (BVP) & 4.75 & 4.75 & 5.00 & 4.75 & 0.75\\ \midrule\midrule
$\mu \pm \sigma$ & $4.875 \pm 0.331$ & $4.625 \pm 0.484$ & $4.625 \pm 0.599$ & $4.500 \pm 0.500$ & $0.688 \pm 1.261$ \\
\bottomrule
\end{tabular}
}
\end{table}
The results indicate that GPT-4o provided highly useful explanations, with average utility scores of 4.875 for analysis, 4.625 for causes, and 4.625 for criticality. While the overall coherence of the explanations was strong, with an average score of 4.5, the low specificity score of 0.688 can be attributed to two main factors. First, as we already remarked, the datasets were collected in a controlled setting, which provided limited contextual information about the patient's activities or other physiological signals. This restricted the ability to contextualize the anomalies with a broader understanding of the patient's behavior or lifestyle. Second, the lack of detailed clinical history for the patient further limited the depth of the explanations. Without access to relevant background information, the model could only offer general insights based on the detected anomalies rather than providing a more personalized interpretation tied to the patient’s specific condition or medical history.
The medical professional was also asked to rate the significance of the anomaly to assess its relevance, as well as its criticality to evaluate the potential risk it poses to the patient, taking into account factors such as age and overall health condition. These ratings were then compared with those provided by GPT-4o to determine whether numerical metrics could be generated to offer an immediately comprehensible score. Such a score would help clinicians rapidly identify anomalies that are of particular concern and may require urgent attention or further investigation.
It should be noted that while GPT-4o generated ratings on a 0-10 scale (see~\autoref{fig:prompt}), the medical professional used a more compact 0-5 scale. To use a finer-grained scale to reduce the effect of small rating errors, we requested GPT-4o to produce results on a 0-10 scale. In this way, we obtain step errors leading to a smaller fractional error when mapped to the 0-5 scale originally used by the doctor. 
Afterward, the GPT-4o ratings were rescaled by dividing by 2, aligning them to the 0-5 range for direct comparison with the doctor’s ratings. 

\begin{figure}[!h]
    \centering
    \includegraphics[width=\linewidth]{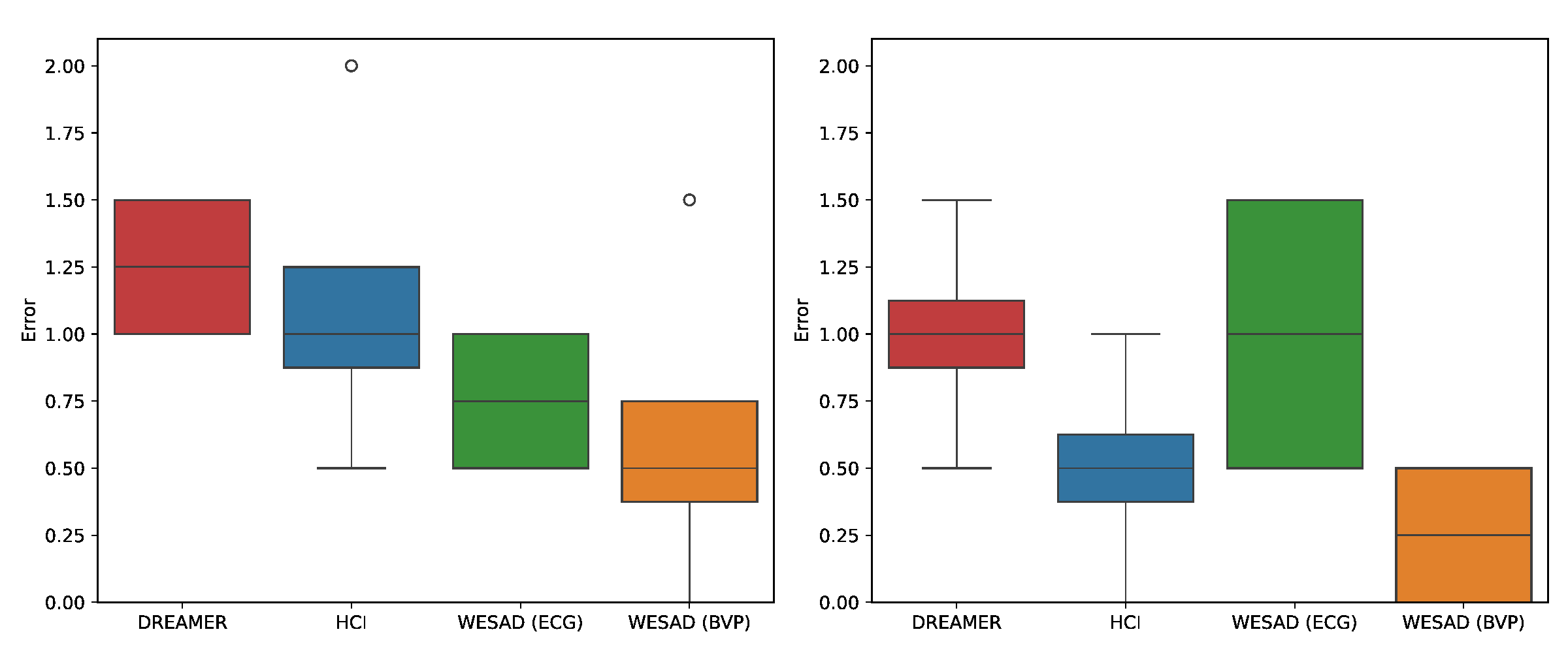}
    \caption{Distribution of errors in GPT-4o's anomaly significance (left) and criticality ratings (right) compared to a medical professional's assessments. Ratings are on a scale from 0 to 5.}
    \label{fig:human-gpt-error}
\end{figure}

The results are presented in~\autoref{tab:significance-criticality} -- for error distributions, refer to~\autoref{fig:human-gpt-error} -- which shows that GPT-4o generally assigned higher ratings across all metrics and datasets. However, the average error rates were 0.938 for significance and 0.688 for criticality, indicating that the model's assessments were comparable to the expert's. These findings suggest that by using an LLM, we can generate useful explanations that support and accelerate the review process of anomalies, providing valuable assistance to medical professionals in assessing and prioritizing potential issues.

\begin{table}[!t]
\centering
\caption{Comparison of anomaly significance and criticality ratings between a medical professional and GPT-4o, showing the errors in GPT estimations for each dataset. The ratings are on a 0-5 scale. Each dataset contains 4 explanations. Here, we show averages for each dataset and the cumulative average ($\mu \pm \sigma$) over all explanation evaluations.}
\label{tab:significance-criticality}
\resizebox{\linewidth}{!}{%
    \begin{tabular}{@{}lcccccc@{}}
    \toprule
    & \multicolumn{3}{c}{\textbf{Significance}} & \multicolumn{3}{c}{\textbf{Criticality}} \\ \cmidrule(l){2-4} \cmidrule(l){5-7} 
      &
      \textbf{\begin{tabular}[c]{@{}c@{}}Human\\ Evaluation\end{tabular}} &
      \textbf{\begin{tabular}[c]{@{}c@{}}GPT\\ Evaluation\end{tabular}} &
      \textbf{Error} &
      \textbf{\begin{tabular}[c]{@{}c@{}}Human\\ Evaluation\end{tabular}} &
      \textbf{\begin{tabular}[c]{@{}c@{}}GPT\\ Evaluation\end{tabular}} &
      \textbf{Error} \\ \midrule
      DREAMER & 1.00 & 2.25 & 1.25 & 0.25 & 1.25 & 1.00\\
      HCI & 1.00 & 2.13 & 1.13 & 0.50 & 1.00 & 0.50\\
      WESAD (ECG) & 1.50 & 2.25 & 0.75 & 0.50 & 1.50 & 1.00\\
      WESAD (BVP) & 1.50 & 2.13 & 0.63 & 0.75 & 1.00 & 0.25 \\\midrule\midrule
$\mu \pm \sigma$ & - & - & $0.938 \pm 0.496 $ & - & - & $0.688 \pm 0.496$\\
\bottomrule
    \end{tabular}
}
\end{table}

\section{Discussion and Conclusion}

\paragraph{Summary of results}
This is the first study to employ a universal time series model, UniTS, to support the seamless monitoring of physiological signals. We conducted an extensive evaluation by benchmarking UniTS against 12 leading methods across three datasets for detecting stress and abnormal emotions.
Our results show that UniTS consistently outperforms state-of-the-art models (see \autoref{quantitative}). Notably, UniTS performs equally well when using both invasive monitoring devices, such as ECG, and more lightweight alternatives like wristbands, demonstrating strong resilience to noisy data. This study also examines how the sampling interval influences the model's performance, demonstrating that robust results can be achieved even with lower-cost monitoring equipment as long as the sampling interval stays within acceptable thresholds.

\paragraph{Benefits in clinical applications}
In the literature, a combination of physiological, behavioral, and psychological monitoring methods have been proposed to monitor the stress level in patients with neurodegenerative diseases,  facilitated by advanced wearable devices and digital technologies ~\cite{reviewstress24}. Speech and video analysis, ECG, EEG, and implantable devices have shown significant promise in stress monitoring for patients with neurodegenerative diseases. These methods have been shown to provide more accurate, continuous, and detailed insights into stress levels than less invasive approaches. However, they present challenges such as higher costs, potential discomfort, privacy concerns, and the need for manual procedures. In this paper, we demonstrated that using last-generation foundational AI models can boost the performance of less invasive devices, making them comparable to those of devices with a greater impact on the daily lives of monitored patients. Monitoring the progression of patients' disease using lightweight devices has numerous advantages in clinical practice, including increased patient comfort and compliance, continuous and real-time monitoring, reduced healthcare costs, increased patient independence, personalized care, improved remote patient management, and improved health outcomes through early intervention.

Furthermore, we have shown how casting the stress detection problem as one of anomaly detection, rather than classification, improves the explainability of the model's outcomes (\autoref{qualitative}). Explainability is vital in medical applications of AI because it ensures trust and accountability, supports clinical validation, and facilitates integration into clinical workflows. In addition, anomaly detection better aligns with the clinical practice since each patient has unique physiological and behavioral baselines. Anomaly detection allows for creating personalized baselines, making identifying deviations that indicate stress easier. 
 

\paragraph{Limitations and future work}
We acknowledge that this study also has limitations. For example, the study compares HR and HRV extracted from a wristband and ECG device, partly due to the wider availability of benchmark datasets for these types of signals. In the future, we plan to extend our experiments to a wider variety of physiological signals that are both detectable by lightweight devices and relevant to studying mental states, such as physical activity and sleep quality. Additionally, we note that the textual explanations of the anomalies, although evaluated as high-quality w.r.t. almost all dimensions considered, have a low specificity because they do not consider the patient's past stress episodes. Currently, the lack of naturalistic datasets does not allow the detection of a mental stress state to be conditional on the patient's past history. Finally, we plan to go beyond stress monitoring to include the observation of other signals of physical and cognitive decline, such as daily activity routines and other actigraphy data, as well as contextual factors. Note that UniTS (see~\autoref{method}) \textit{by design} allows for the integration of time series of different types and lengths. Similar to the limitations mentioned above, the obstacle to this type of experiment is rather the absence of datasets that collect, for each monitored patient, multiple signals from multiple devices in naturalistic settings.  

To overcome these limitations, we are currently collecting, in the context of the Regional project @HOME, a variety of physiological, behavioral, and environmental signals for a cohort of 10 patients monitored in their homes\footnote{We are aware of the privacy and ethic concerns implied by this experiment. However, discussing how to mitigate privacy and other risk factors is outside the scope of this paper.}. We expect that the collection of these data will help to demonstrate more clearly the advantages of the approach presented in this paper and also constitute a valuable benchmark for other researchers.

\section*{Acknowledgements}
\noindent This work was supported by a grant from Lazio Region, FESR Lazio 2021-2027, project @HOME (\# F89J23001050007
CUP B83C23006240002). We sincerely thank Prof. Dr. Emilia Reda for her valuable contribution to evaluating the quality of the diagnoses provided by the system described in this article.

\bibliographystyle{plain} 
\bibliography{bibliography}

\begin{appendix}
    
\end{appendix}
\section{Dataset characteristics}\label{app:datasets}
 \autoref{tab:dataset-characteristics} summarizes the key characteristics of this study's DREAMER, HCI, and WESAD datasets. The values reported are averaged over 5 cross-validation folds. 
We outline important data attributes such as the number of training and test instances, class imbalance and anomaly ratios, and the mean and standard deviation of the HR and HRV features.

\begin{table}[!t]
\caption{Dataset characteristics for DREAMER, HCI, and WESAD (ECG and BVP) with train and test instance counts, class imbalance, anomaly ratios, and descriptive statistics for feature distributions.}
\label{tab:dataset-characteristics}
\resizebox{\linewidth}{!}{%
    \begin{tabular}{@{}lcccc@{}}
    \toprule
  & \textbf{DREAMER} & \textbf{HCI}    & \textbf{WESAD (ECG)} & \textbf{WESAD (BVP)} \\
    \midrule
    \# of Train Instances         & 8297         & 6172         & 6088          & 6088         \\
    \# of Test Instances          & 3404         & 1966         & 1720          & 1720         \\
    \# Anomalies (Test)           & 1330         & 423          & 199           & 199          \\
    \midrule
    Imbalance Ratio         & 0.654          & 0.276          & 0.131           & 0.131          \\
    Anomaly Ratio                 & 0.391          & 0.215          & 0.115           & 0.115          \\
    \midrule
    HR Feature (Train)  & 69.71 ± 11.16 & 74.21 ± 14.93 & 75.14 ± 12.48 & 77.14 ± 11.52 \\
    HR Feature (Test)  & 69.69 ± 10.75  & 74.33 ± 15.08 & 77.84 ± 13.44      & 81.37 ± 13.54      \\
    HRV Feature (Train) & 13.07 ± 8.31  & 35.80 ± 62.28 & 42.65 ± 30.45 & 38.08 ± 24.67 \\
    HRV Feature (Test) & 13.12 ± 8.03   & 35.98 ± 57.54  & 41.57 ± 26.04      & 38.46 ± 23.06      \\
    \bottomrule
    \end{tabular}
}
\end{table}

\section{Reproducibility guidelines}\label{app:reproducibility}
To ensure the reproducibility of the experiments conducted in this study, the following information regarding the datasets, software, and hardware used is provided. Our code is publicly available at \url{https://github.com/davegabe/Wearable-Stress-Monitor}.

\paragraph{Dataset Information}
This study utilizes datasets originally designed for stress detection and emotion classification, repurposed for anomaly detection tasks, as detailed in \autoref{sec:datasets}. HR and HRV features were extracted from raw ECG and BVP signals (for WESAD).

The preprocessing pipeline for the ECG and BVP signals follows several key steps. First, the raw signals are filtered to remove noise and ensure they fall within the relevant frequency range. A bandpass filter between 3-45 Hz is applied to the ECG signals to isolate heart-related frequencies, while a bandpass filter between 1-8 Hz is applied to the BVP signals to focus on pulse wave frequencies. After filtering, R-peak detection is performed on the ECG signal to identify the location of heartbeats using algorithms such as the Hamilton segmenter. For the BVP signal, onset detection methods are applied to identify the start of each pulse wave.

The time intervals between successive R-peaks (the RR intervals for ECG) and pulse onsets (for BVP) are used to compute HR. These intervals are converted into heart rate in beats per minute (bpm). HRV analysis uses RMSSD (Root Mean Square of Successive Differences), a common time-domain metric calculated from the RR or pulse intervals. The preprocessing pipeline is applied over 60-second windows with a 10-second sliding window, simulating data capture from wearable devices such as smartwatches.

\paragraph{Software and Technologies}
This study utilized a range of Python libraries and code repositories to implement and evaluate various anomaly detection models. Specifically, \texttt{biosppy}~\cite{biosppy} was used to process ECG and BVP signals through a pipeline involving signal filtering, R-peak detection, and HR computation, while \texttt{pyhrv}~\cite{Gomes2019} was employed to compute HRV.

The \textit{TranAD repository} proved invaluable, providing implementations of a wide range of state-of-the-art models, including TranAD, LSTM-NDT, CAE-M, DAGMM, USAD, OmniAnomaly, MTAD-GAT, GDN, MAD-GAN, and MSCRED, allowing for a comprehensive evaluation of diverse approaches. We also utilized the \textit{HypAD repository} for implementing HypAD and TadGAN, while the \textit{UniTS repository} supplied both the code and pre-trained weights for the foundational UniTS model.

The choice of these repositories was driven by their extensive coverage of anomaly detection models, which facilitated a diverse and comprehensive evaluation of different techniques. This also allowed us to build on existing, well-documented implementations, ensuring the reproducibility of our experiments.

\paragraph{Hardware Specifications}
The experiments were conducted using a machine equipped with an Intel Xeon E5-2650 v4 CPU, 32 GB of RAM, and an NVIDIA RTX 3060 GPU.

\section{Extended qualitative analysis}\label{app:qualitative}

\paragraph{Reconstruction errors confirm that UniTS pays attention to multivariate signal interdependencies for more accurate anomaly detection}

\begin{sidewaysfigure}[!ht]
    \centering
    \includegraphics[width=\textwidth]{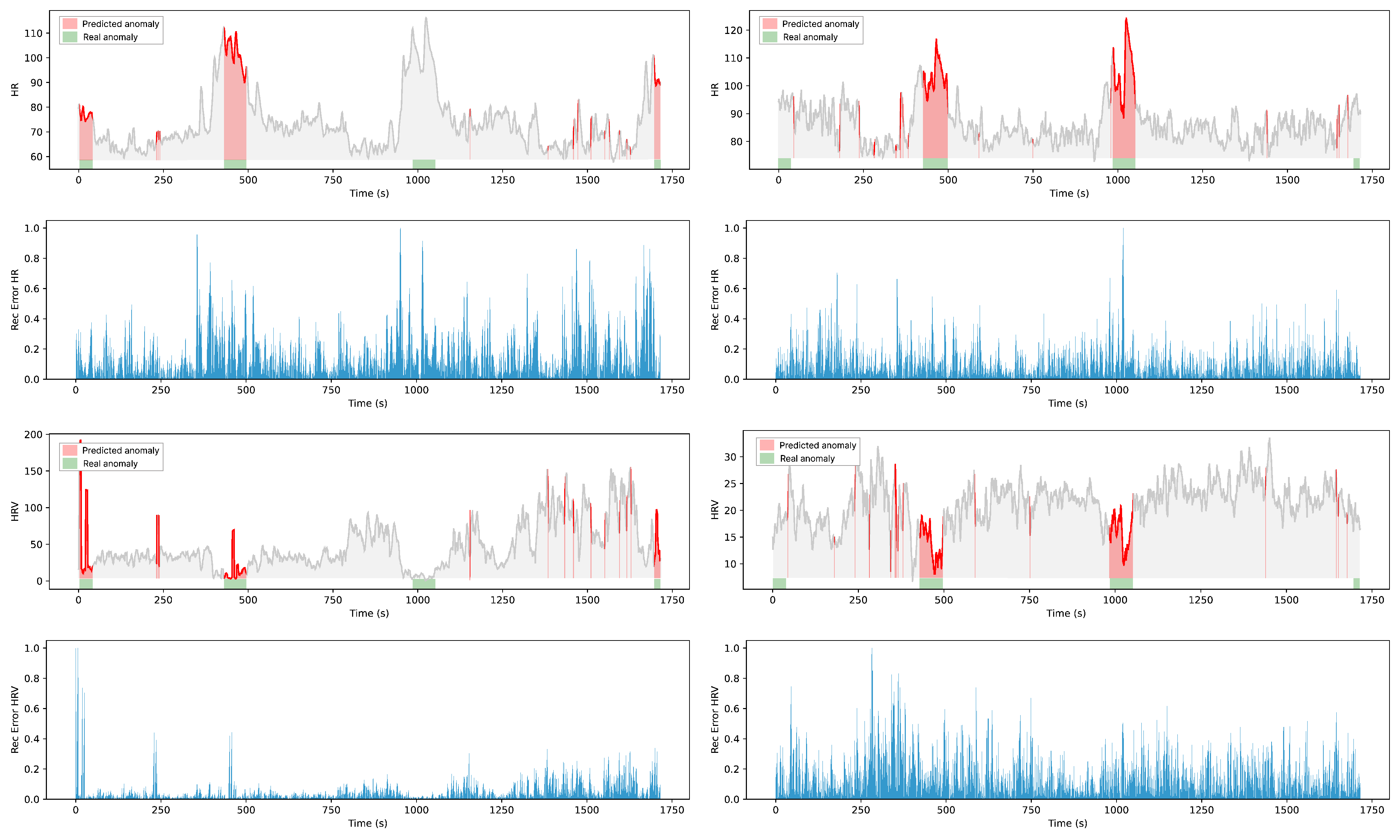}
    \caption{Anomaly predictions on WESAD ECG (left) vs WESAD BVP (right) according to HR and HRV with their respective reconstruction errors. The top row in each column shows the HR trends over time, with predicted anomalies highlighted in red and real anomalies in green. The second row displays the reconstruction error for HR. The third row shows the HRV trends with similar anomaly markings, and the final row plots the reconstruction error for HRV. }
    \label{fig:qualitative_hr_vs_hrv}
\end{sidewaysfigure}

\autoref{fig:qualitative_hr_vs_hrv} compares anomaly detection in HR and HRV from the WESAD ECG (left) and WESAD BVP (right) datasets. The top two rows display HR values and reconstruction errors for each dataset, while the bottom two rows show HRV values and the corresponding reconstruction errors. In both columns, predicted anomalies (in red) and real anomalies (in green) are highlighted, offering a clear visual of the model’s performance. There are instances in both ECG and BVP signals where predicted anomalies align well with real anomalies, though several mismatches also occur, indicating the presence of false positives and false negatives. The reconstruction errors for both HR and HRV are more pronounced at certain times, suggesting moments of heightened physiological variability or noise in the data, with notable peaks in the error plots that do not always correspond to predicted anomalies. Interestingly, HRV signals in WESAD BVP exhibit greater fluctuations than those in the WESAD ECG, possibly reflecting differences in how these signals capture physiological variability. The comparison shows that UniTS may perform differently across the two data sources, with a tendency to detect more clustered HRV anomalies in WESAD BVP. This highlights the potential need for adjustments in anomaly detection methods based on signal type. Moreover, in WESAD ECG, the first anomaly occurrence is a clear example of how the reconstruction error for HR helps UniTS accurately identify an anomaly that aligns with the ground truth. A noticeable HR reconstruction error (second row) spike coincides with the model’s predicted anomaly. However, the reconstruction error for HRV at the same time point does not show a significant spike, suggesting that the model’s ability to detect anomalies relies on the multivariate signal dependency, where one signal’s error compensates for the lack of a response in the other.

\begin{figure}[!h]
    \begin{boxA}
        \scriptsize
        \textcolor{black}{
            \vspace{-1em}
            \begin{center}
                \includegraphics[width=0.549\textwidth]{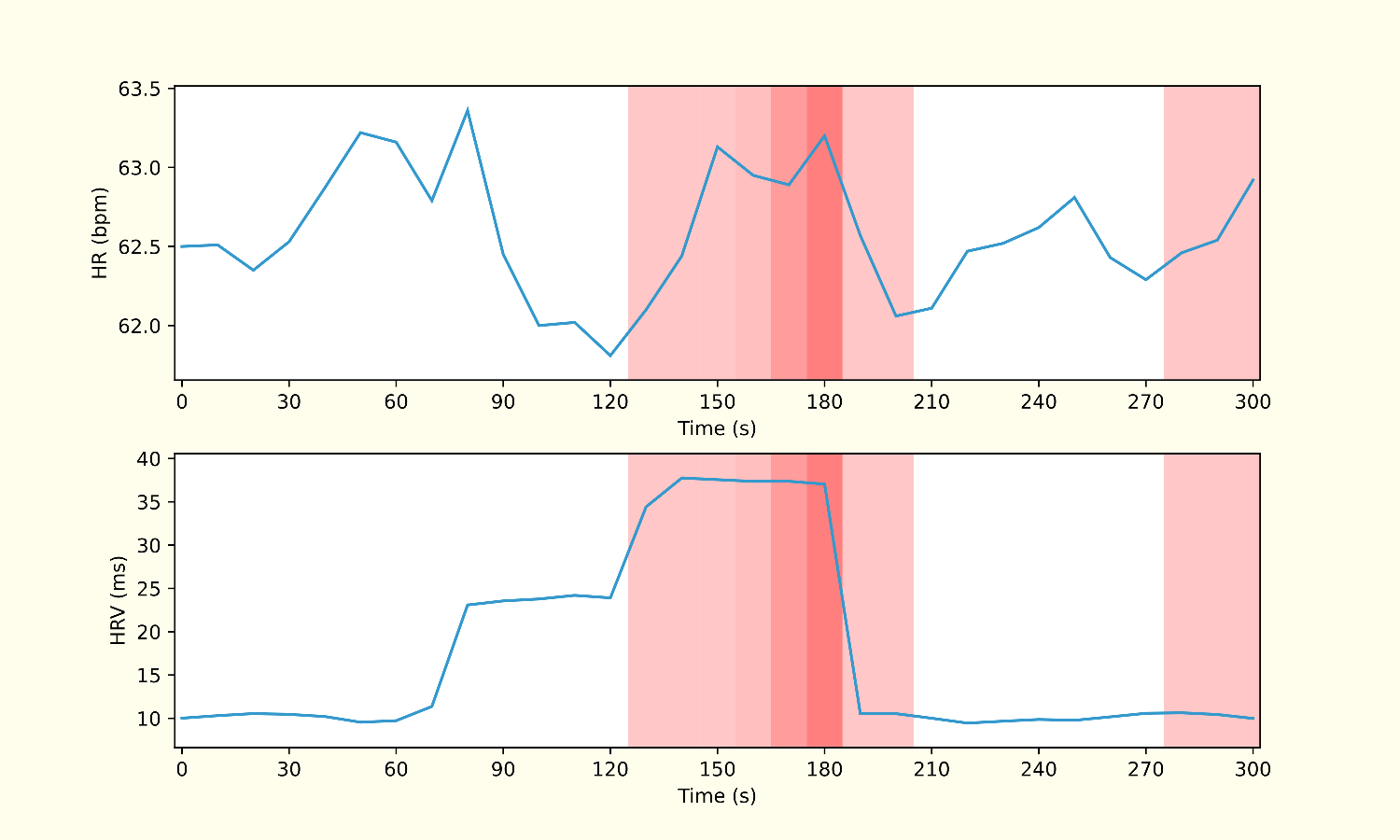}
            \end{center}
                \vspace{-0.5em}
                \textbf{\colorbox{mygreen}{\textcolor{white}{Analysis}}}
                \vspace{-0.8em}
                \begin{itemize}
                    \setlength{\itemsep}{-3pt}
                    \item The data shows a notable increase in HRV from 23.91 ms at 120 seconds to 37.75 ms at 140 seconds, coinciding with the detection of anomalies.
                    \item A series of high-intensity anomalies are detected between 130 and 180 seconds, with HRV peaking at 37.75 ms and HR remaining relatively stable around 62-63 bpm.
                    \item After 180 seconds, HRV sharply decreases to baseline levels, and anomalies cease until a brief reoccurrence at 280 seconds.
                    \item The HR remains relatively stable throughout the recording, with minor fluctuations, suggesting changes in HRV primarily drive the anomalies.
                \end{itemize}
                \vspace{-0.5em}
                \textbf{\colorbox{myblue}{\textcolor{white}{Causes}}}
                \vspace{-0.8em}
                \begin{itemize}
                    \setlength{\itemsep}{-3pt}
                    \item The sudden increase in HRV during the anomaly period could indicate an emotional or physiological response, such as stress or excitement, common in young adults.
                    \item The anomalies could also be due to measurement artifacts or environmental factors affecting the sensor, though the consistent pattern suggests a physiological basis.
                    \item Given the patient's demographic and lack of medical conditions, the anomalies are less likely due to underlying health issues.
                \end{itemize}
                \vspace{-0.5em}
                \textbf{\colorbox{myred}{\textcolor{white}{Criticality}}}
                \vspace{-0.8em}
                \begin{itemize}
                    \setlength{\itemsep}{-3pt}
                    \item The high-intensity anomalies observed between 130 and 180 seconds are significant and warrant attention, as they could indicate a strong emotional response.
                    \item While the anomalies are not immediately critical, they should be monitored to ensure they do not recur frequently or increase in intensity.
                    \item The brief anomaly at 280 seconds appears less significant and may not require immediate follow-up unless it becomes more frequent.
                \end{itemize}
                \vspace{-0.5em}
                \textbf{\colorbox{myorange}{\textcolor{white}{Scores}}}
                \vspace{-0.8em}
                \begin{itemize}
                    \setlength{\itemsep}{-3pt}
                    \item \textbf{Significance Score: 7/10} - The anomalies are clinically relevant, indicating a potential emotional response, but not suggestive of a critical health issue.
                    \item \textbf{Criticality Score: 4/10} - The anomalies are not urgent but should be monitored to prevent potential escalation or recurrence.
                \end{itemize}
        }
    \end{boxA}
    \caption{Explanation of anomalies detected in sample data from the DREAMER dataset, generated by GPT-4o. The explanation is organized into four sections (Analysis, Causes, Criticality, and Scores) providing a structured evaluation of the detected anomalies and their potential implications. Notice that the plot with the monitored signals contains red bands corresponding to the detected anomalies. The more intense the color, the more acute the anomaly detected (in terms of anomaly energy).    }
\label{fig:dreamer-example}
\end{figure}

\begin{figure}[!h]
    \begin{boxA}
        \scriptsize
        \textcolor{black}{
            \vspace{-1em}
            \begin{center}
                \includegraphics[width=.549\textwidth]{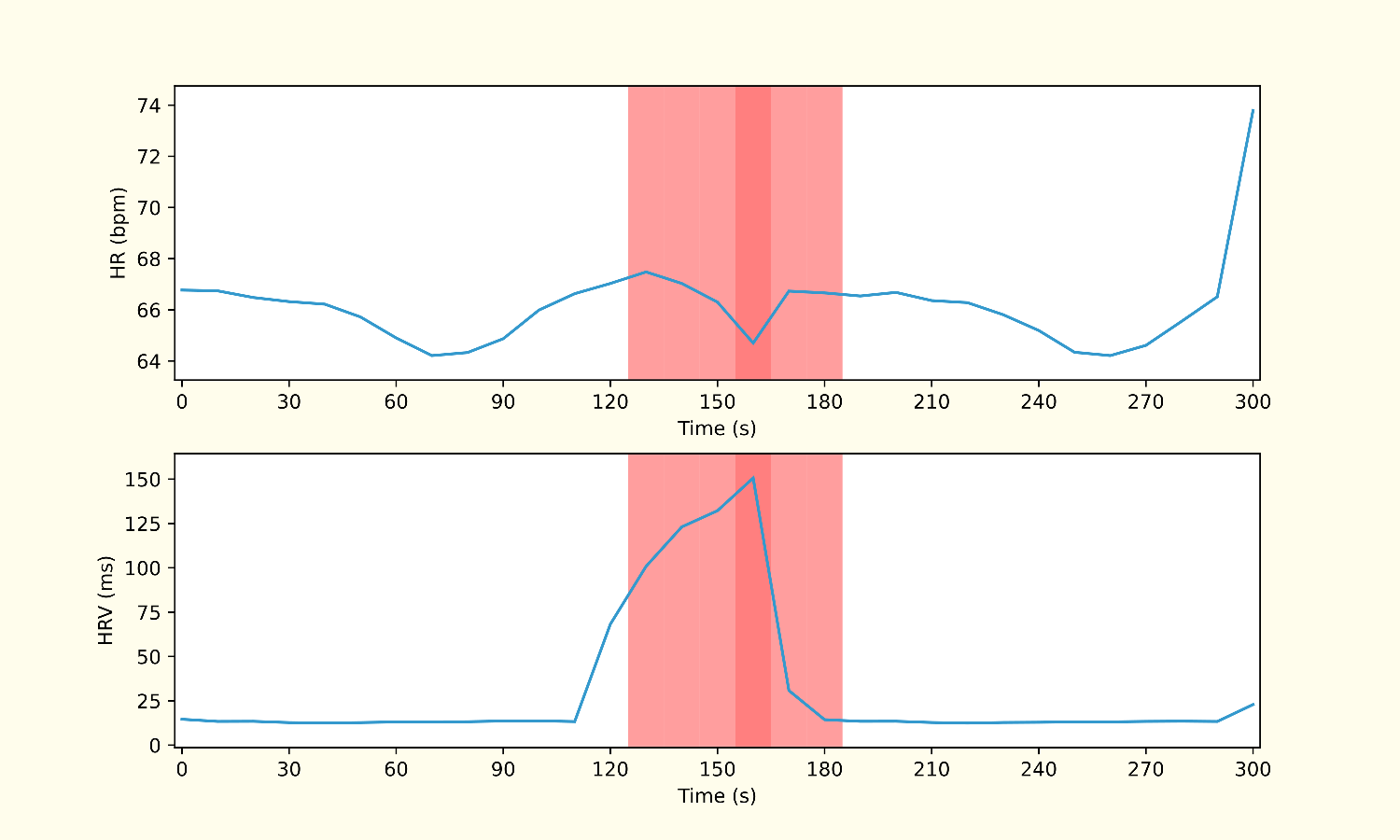}
            \end{center}
                \vspace{-0.5em}
                \textbf{\colorbox{mygreen}{\textcolor{white}{Analysis}}}
                    \vspace{-0.8em}
                    \begin{itemize}
                    \setlength{\itemsep}{-3pt}
                    \item A significant anomaly is detected between 130 and 180 seconds, characterized by a marked increase in HRV, peaking at 150.64 ms, while HR remains relatively stable around 66-67 bpm. This suggests a strong physiological response.
                    \item Post-anomaly, HRV sharply decreases to baseline levels, indicating a return to normal physiological state.
                    \item A secondary, less pronounced anomaly is observed at 300 seconds, with a sudden HR increase to 73.79 bpm and a moderate HRV of 22.9 ms, suggesting a transient physiological change.
                \end{itemize}
                \vspace{-0.5em}
                \textbf{\colorbox{myblue}{\textcolor{white}{Causes}}}
                \vspace{-0.8em}
                \begin{itemize}
                    \setlength{\itemsep}{-3pt}
                    \item The primary anomaly could be due to an emotional response, such as stress or excitement, given the patient's young age and absence of medical conditions. The sudden increase in HRV suggests a strong autonomic nervous system response.
                    \item The secondary anomaly at 300 seconds might be due to a brief physical activity or environmental factor, such as a change in posture or external stimulus.
                    \item System errors or data noise are unlikely given the clear pattern and return to baseline, but cannot be completely ruled out.
                \end{itemize}
                \vspace{-0.5em}
                \textbf{\colorbox{myred}{\textcolor{white}{Criticality}}}
                \vspace{-0.8em}
                \begin{itemize}
                    \setlength{\itemsep}{-3pt}
                    \item The primary anomaly is clinically significant due to its intensity and duration, warranting monitoring to rule out any underlying issues or recurrent stressors.
                    \item The secondary anomaly is less critical but should be noted for future patterns.
                    \item Immediate follow-up is unnecessary unless similar patterns recur frequently, suggesting a persistent stressor or physiological issue.
                \end{itemize}
                \vspace{-0.5em}
                \textbf{\colorbox{myorange}{\textcolor{white}{Scores}}}
                \vspace{-0.8em}
                \begin{itemize}
                    \setlength{\itemsep}{-3pt}
                    \item \textbf{Significance Score: 7/10} - The primary anomaly is significant due to its intensity and potential link to emotional or physiological stress.
                    \item \textbf{Criticality Score: 4/10} - While notable, the anomalies do not pose an immediate threat but should be monitored for recurrence.
                \end{itemize}
        }
    \end{boxA}
    \caption{Explanation of anomalies detected in sample data from the HCI dataset, generated by GPT-4o. The explanation is organized into four sections (Analysis, Causes, Criticality, and Scores) providing a structured evaluation of the detected anomalies and their potential implications. Notice that the plot with the monitored signals contains red bands corresponding to the detected anomalies. The more intense the color, the more acute the anomaly detected (in terms of anomaly energy).}
\label{fig:hci-example}
\end{figure}

As discussed in~\autoref{par:zeroshotllm},~\autoref{fig:dreamer-example} and~\autoref{fig:hci-example} illustrate two examples of the zero-shot qualitative evaluation regarding analysis, causes, criticality, and scores from GTP-4o beginning from a plot containing the monitored signals and detected anomalies (see~\autoref{fig:prompt} for an example prompt given to GPT-4o to generate explanations). Supported by~\autoref{tab:comparison},~\autoref{tab:significance-criticality} and~\autoref{fig:human-gpt-error}, one can use the significance and criticality scores to compare with expert evaluators and assess the usefulness and adaptability of LLMs in healthcare domains.

\end{document}